\newif\ifdraft
\draftfalse
\PassOptionsToPackage{unicode}{hyperref}
\PassOptionsToPackage{hyphens}{url}

\documentclass[
]{article}
\usepackage{tabularx}
\usepackage{booktabs} % For nice-looking tables
\usepackage{array} % For better column formatting
\usepackage{caption} % For captions
\usepackage{afterpage}
\usepackage{amsmath,amssymb}
\usepackage[normalem]{ulem} % provides \sout
\usepackage{xcolor}         % for \textcolor

\usepackage{iftex}
\ifPDFTeX
  \usepackage[T1]{fontenc}
  \usepackage[utf8]{inputenc}
  \usepackage{textcomp} % provide euro and other symbols
\else % if luatex or xetex
  \usepackage{unicode-math} % this also loads fontspec
  \defaultfontfeatures{Scale=MatchLowercase}
  \defaultfontfeatures[\rmfamily]{Ligatures=TeX,Scale=1}
\fi
\usepackage{lmodern}
\ifPDFTeX\else
\fi
\newcommand{\keywords}[1]{{\small\textbf{Keywords:} #1}}

\IfFileExists{upquote.sty}{\usepackage{upquote}}{}
\IfFileExists{microtype.sty}{% use microtype if available
  \usepackage[]{microtype}
  \UseMicrotypeSet[protrusion]{basicmath} 
}{}
\makeatletter
\@ifundefined{KOMAClassName}{% if non-KOMA class
  \IfFileExists{parskip.sty}{%
    \usepackage{parskip}
  }{% else
    \setlength{\parindent}{0pt}
    \setlength{\parskip}{6pt plus 2pt minus 1pt}}
}{% if KOMA class
  \KOMAoptions{parskip=half}}
\makeatother
\usepackage{authblk}
\usepackage{xcolor}
\usepackage{longtable,booktabs,array}
\usepackage{multirow}
\usepackage{pdfpages}
\usepackage{calc} % for calculating minipage widths
\usepackage{etoolbox}
\makeatletter
\patchcmd\longtable{\par}{\if@noskipsec\mbox{}\fi\par}{}{}
\makeatother
\IfFileExists{footnotehyper.sty}{\usepackage{footnotehyper}}{\usepackage{footnote}}
\makesavenoteenv{longtable}
\usepackage{graphicx}
\makeatletter
\def\maxwidth{\ifdim\Gin@nat@width>\linewidth\linewidth\else\Gin@nat@width\fi}
\def\maxheight{\ifdim\Gin@nat@height>\textheight\textheight\else\Gin@nat@height\fi}
\makeatother
\setkeys{Gin}{width=\maxwidth,height=\maxheight,keepaspectratio}
\makeatletter
\def\fps@figure{htbp}
\makeatother
\ifLuaTeX
  \usepackage{luacolor}
  \usepackage[soul]{lua-ul}
\else
  \usepackage{soul}
\fi
\ifLuaTeX
  \usepackage{selnolig}  % disable illegal ligatures
\fi
\IfFileExists{bookmark.sty}{\usepackage{bookmark}}{\usepackage{hyperref}}
\IfFileExists{xurl.sty}{\usepackage{xurl}}{} % add URL line breaks if available
\hypersetup{
  hidelinks,
  pdfcreator={LaTeX via pandoc}}

\ifdraft
  \newcommand{\added}[1]{\textcolor{blue}{#1}}
  \newcommand{\deleted}[1]{\textcolor{red}{\sout{#1}}}
  \newcommand{\replaced}[2]{\textcolor{blue}{#1} \textcolor{red}{\sout{#2}}}
  \newcommand{\deletedfloat}[1]{}%{\textcolor{red}{#1}}
  \newcommand{\commented}[1]{\textcolor{blue}{#1}}
\else
  \newcommand{\added}[1]{#1}
  \newcommand{\deleted}[1]{}
  \newcommand{\replaced}[2]{#1}
  \newcommand{\deletedfloat}[1]{}
  \newcommand{\commented}[1]{}
\fi

\title{Machine-learning for photoplethysmography analysis: Benchmarking
feature, image, and signal-based approaches}

\author[1]{Mohammad Moulaeifard}
\author[2]{Loic Coquelin}
\author[3]{Mantas Rinkevičius}
\author[3]{Andrius Sološenko}
\author[4]{Oskar Pfeffer}
\author[5]{Ciaran Bench}
\author[4]{Nando Hegemann}
\author[6]{Sara Vardanega}
\author[6]{Manasi Nandi}
\author[6]{Jordi Alastruey}
\author[7]{Christian Heiss}
\author[3]{Vaidotas Marozas}
\author[5]{Andrew Thompson}
\author[5,8]{Philip J. Aston}
\author[9]{Peter H. Charlton}
\author[1]{Nils Strodthoff}

\affil[1]{Carl von Ossietzky Universität Oldenburg, Oldenburg, Germany}
\affil[2]{Laboratoire national de métrologie et d'essais, Paris, France}
\affil[3]{Biomedical Engineering Institute, Kaunas University of Technology, Kaunas, Lithuania}
\affil[4]{Physikalisch-Technische Bundesanstalt, Berlin, Germany}
\affil[5]{National Physical Laboratory, Teddington, United Kingdom}
\affil[6]{King's College London, London, United Kingdom}
\affil[7]{University of Surrey, Surrey, Guildford, United Kingdom}
\affil[8]{School of Mathematics and Physics, University of Surrey, Guildford, United Kingdom}
\affil[9]{Department of Public Health and Primary Care, University of Cambridge, Cambridge, United Kingdom}
\date{}
\begin{document}

% Insert external PDF as first page
%\includepdf[pages=1]{D1_cover_sheet.pdf}
\maketitle

\begin{abstract}

\textbf{Background:} Photoplethysmography (PPG) is a non-invasive physiological sensing method used in many clinical applications, increasingly supported by machine learning. However, systematic comparisons of input representations and models remain limited. 
\textbf{Methods:} We address this gap in the research landscape by a comprehensive benchmarking study covering three kinds of input representations, interpretable features, image representations and raw waveforms, across prototypical regression and classification use cases: blood pressure (BP) estimation and atrial fibrillation (AF) prediction.
\textbf{Results:} \added{In both cases, the best results are achieved by deep neural networks operating on raw time series as input representations. Within this model class, the strongest performance is observed for deeper convolutional neural networks (CNNs). However, depending on the task, smaller or lower-capacity CNNs can  also achieve competitive performance, as confirmed by Bland–Altman analyses and statistical significance analyses based on bootstrapping.}
\textbf{Conclusions:} \added{By providing a controlled, like-for-like comparison across signal, feature, and image-based representations,  this study offers practical guidance for selecting robust machine-learning approaches for real-world PPG applications.} 
\end{abstract}


\hypertarget{introduction}{%
\section{Introduction}\label{introduction}}

PPG is one of the most commonly used wearable sensing techniques. It
consists of projecting light onto the skin and measuring the amount of
light that is reflected back or transmitted through the underlying
tissues. Its simplicity, non-invasive nature, and ability to deliver
multiple physiological parameters make it particularly attractive \cite{ref1,ref2}.
As a result, PPG has been integrated into a range of clinical devices,
such as pulse oximeters, as well as consumer wearable devices, including smartwatches.

\begin{figure}[t]  
  \centering
  \includegraphics[width=\columnwidth]{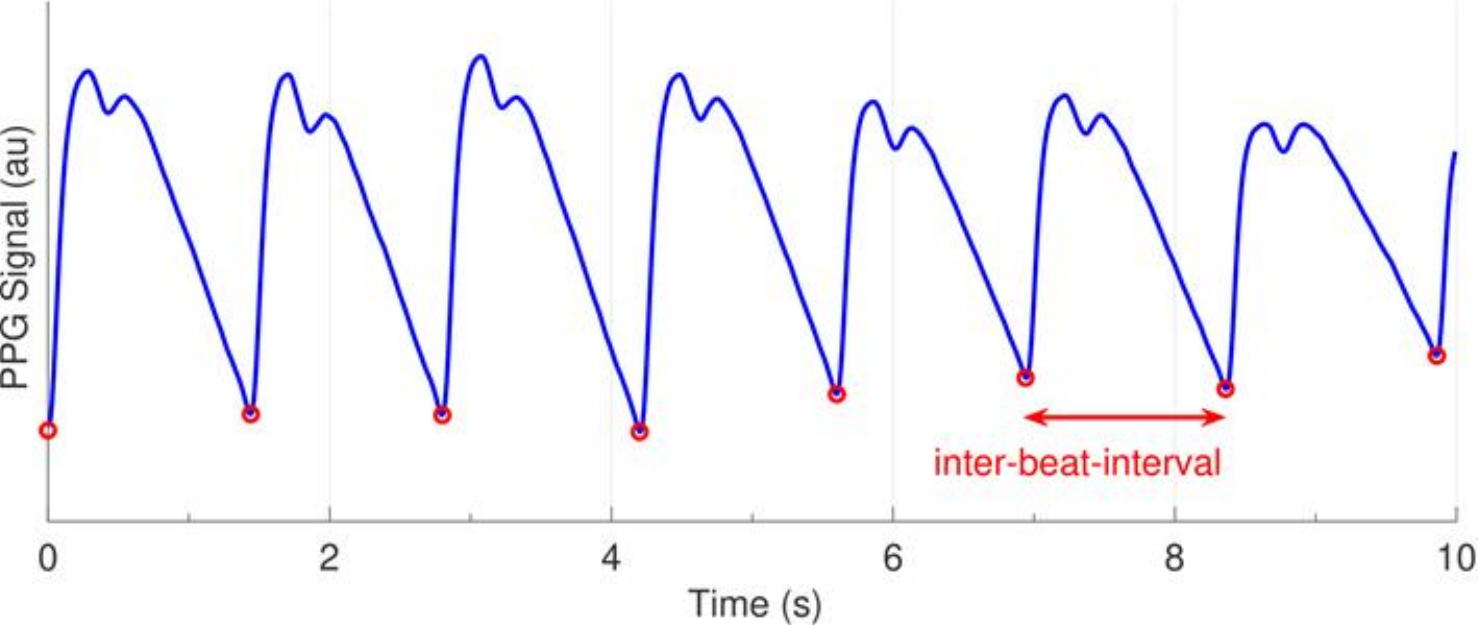}
  \caption{An exemplary PPG signal showing a pulse wave for each heartbeat. Pulse onsets, representing individual heartbeats, are shown as red circles. An inter-beat interval is labeled, corresponding to the time between consecutive heartbeats (adapted from \cite{ref7}).} 
  \label{fig:fig1}
\end{figure}

The PPG signal measures the fluctuations in blood volume in the skin's
microvascular bed, which occur with each heartbeat. Figure \ref{fig:fig1} shows an
exemplary PPG signal: the shape of the pulse wave contains information
relating to the heart and vasculature, including BP, and the
inter-beat intervals are related to heart rhythm \cite{ref3}. The time delay
between the electrical activation of the heart and the arrival of the
corresponding pulse wave at a peripheral site where PPG is measured has
been used to predict BP. There are generally two approaches
to analysing PPG signals \cite{ref1}: (i) using signal processing to extract
features relating to pulse wave shape or inter-beat-intervals; and (ii)
using deep learning techniques to analyse PPG signals or their
image-based representations. Over the past few years, deep learning
techniques have become widely used \cite{ref5,ref6,nie2024review}, as they have often shown superior performance to feature-based approaches.

\added{In the present study}, we investigate two widely considered clinical applications
for PPG analysis: AF classification as a
prototypical classification task and cuffless BP
estimation as a prototypical regression task. AF is the most common
sustained cardiac arrhythmia and confers a five-fold increase in stroke
risk \cite{ref8}. AF is characterised by irregular and often very rapid heart
rhythm. PPG provides an attractive approach to identifying AF because it
is widely used in consumer wearables, and because it can detect each
heartbeat, it can provide measures of the heart rhythm. During an AF episode,
the time intervals between heartbeats are irregular. BP is one of the
most widely used physiological measurements. It is a key marker of
cardiovascular health; a valuable predictor of cardiovascular events; and is essential for the selection and monitoring of antihypertensive
(BP lowering) treatments \cite{ref9}. PPG-based BP estimation
provides a potential approach to monitor BP unobtrusively in daily life. \added{Accordingly, numerous studies have explored non-invasive and cuffless BP monitoring using PPG and related physiological signals, spanning feature-based methods as well as deep learning approaches \cite{ref6,nie2024review}.}
The rationale behind selecting two prototypical, but very different use
cases, is to identify general patterns that could guide practitioners in
the field even beyond the two investigated use cases.

While many prediction models have been put forward for the two
considered prediction tasks at hand, benchmarking results are typically
presented within a set of prediction models operating on a single kind
of input modality, or a comparison is carried out against previously
reported results from the literature. However, the latter relies on
matching the experimental setup as closely as possible, where deviations
from this setup severely limits the comparability of the results. \added{The
current lack of directly comparable benchmarking is an important gap in the
research landscape, which we aim to address with this submission. This lack of controlled, directly comparable benchmarking across input
representations and model families motivates the present study.}

In this work, we address the following research questions: How do
state-of-the-art machine learning models operating on different inputs
representations compare? Are there universal patterns in terms of
best-performing input representations or model architectures across
different prototypical classification and regression use cases? \added{As our
main technical contribution, we put forward a like-for-like benchmarking
of a comprehensive set of state-of-the-art algorithms covering three
kinds of input representations on two large-scale datasets for a
prototypical classification task (AF classification) and two different
variants of a prototypical regression task (BP regression), using identical data splits and evaluation protocols to enable direct comparison across model families
and to identify robust performance patterns that generalize beyond a single
model or dataset.}

\hypertarget{related-work}{%
\section{Related Work}\label{related-work}}

\textbf{\added{Overview of Learning Paradigms for PPG Analysis.}} \added{To clarify the structure of this section, we organize prior work primarily by the input representation used for PPG analysis rather than by the learning algorithm. Specifically, we distinguish between models operating on raw time-series signals, engineered or clinically interpretable features, and image-based representations. While similar architectures (e.g., CNNs or RNNs) may appear across categories, the defining criterion is the form of the input representation.} Typically, all approaches for clinical prediction tasks based on PPG data rely on a combination of
signal processing and machine learning, where the precise focus varies
considerably across different approaches. At one end of the spectrum,
approaches relying on clinically interpretable features focus heavily on
signal processing to extract meaningful features, and typically use
comparably simple classification/regression models to perform the
prediction. At the other end of the spectrum, deep learning methods
using raw time series as input with as little signal processing as possible, rely on complex prediction models to extract and process
meaningful features by themselves. In between, there are classifiers
based on image-representations, that leverage signal processing tools to
turn raw waveforms into image representations, and then most commonly
also rely on deep learning models to perform the prediction.

\textbf{\added{Raw Time Series as Input Representations.}} Recent advancements in deep learning have
significantly impacted healthcare by enabling complicated analysis of
raw physiological data. These models are particularly effective in
estimating BP and detecting AF, among other applications. The key
advantage of deep learning is its ability to recognize complicated
patterns directly from raw data such as electrocardiogram (ECG) and PPG
signals, eliminating the need for extensive manual feature development
\cite{ref10,ref11, ref61}. They provide an effective means to learn the complex,
nonlinear underlying relationship between PPG signals and various
physiological parameters, without the need to define a convenient
analytical form for such a transformation. Consequently, deep learning
architectures such as convolutional neural networks (CNNs) and recurrent
neural networks (RNNs) have shown remarkable performance in BP
estimation, by capturing the temporal and spatial nuances inherent in
raw physiological signals \cite{ref12}. Very recent work also demonstrated strong performance of hybrid CNN–BiLSTM architectures with attention on large datasets \cite{mohammadi2025cuff}.

Similarly, deep learning models have
demonstrated significant potential in detecting AF from raw ECG signals.
The authors of \cite{ref11} developed a deep learning model employing a CNN to
detect AF from single-lead ECG recordings. Their model demonstrated high
accuracy, underscoring the potential of deep learning in the detection
of arrhythmias. Similarly, \cite{ref14,cheng2020atrial,ref16} implemented deep learning approaches
using CNNs on PPG signals for AF detection. These approaches yielded
results that were competitive with ECG-based methods, thereby
demonstrating the feasibility of utilizing PPG signals for AF detection.
End-to-end deep learning frameworks that process raw ECG data to
generate AF predictions have simplified and improved the accuracy of AF
detection.

Beyond PPG-specific investigations, several prior works \cite{ismail2020inceptiontime,ref39} have put forward benchmarks on a diverse set of time series classification tasks, establishing, for example, InceptionTime and MiniRocket as robust baselines models, which encourage their use in PPG-based prediction tasks.

\added{More recently, transformer-based architectures have been explored for PPG and other physiological time-series analysis. Several studies have reported competitive performance of Transformer or hybrid CNN–Transformer models for time series forecasting \cite{wu2022timesnet, liu2023itransformer}. However, their effectiveness relative to convolutional architectures remains task- and data-dependent \cite{eldele2024tslanet}. }

\textbf{\added{Feature-Based Input Representations.}} Besides working on raw data, another
possibility to solve classification or regression tasks on PPG data is
to establish machine learning models operating on clinically
interpretable PPG features. This includes features based on
pulse morphology, e.g., \ PPG pulse wave features such as the systolic
peak, the diastolic peak or pulse duration, and PPG derivative features \cite{ref6,ref17}, and irregularity features based on measures of randomness,
variability and complexity in the inter-beat-intervals that can be
determined from the PPG \cite{ref18,ref19}. These clinically interpretable PPG
features are used for BP estimation (see below), AF detection (see
below), and other questions related to the cardiovascular system, e.g.\
the assessment of arterial stiffness \cite{ref20}. In addition, PPG signals sometimes show a strict periodic behaviour or, in general, a quasi-periodic
behaviour. This motivates the use of features from Fourier- \cite{ref21},
Wavelet- \cite{ref22} or Hilbert-Huang- \cite{ref23} Transformations for training
models. Such models are defined for BP estimation and AF detection. In
related work, features from Fourier-Transformation are used to define
models for the detection of aneurysms \cite{ref24} or stenoses \cite{ref25}. As PPG
signals are in general not periodic, but quasiperiodic, one might expect
better results with Wavelet- or Hilbert-Huang-transformations.

\textbf{\added{Image-Based Input Representations.}} Image-based models, such as those using the
Continuous Wavelet Transform (CWT) \added{or short-time Fourier transform (STFT)}, convert physiological signals into
visual representations, enabling deep learning to analyze, classify, and
estimate physiological parameters. 

The CWT is a powerful tool for
analyzing localized variations of power within a time series signal.
Unlike the Fourier Transform, which provides a global frequency
representation, CWT can provide a time-frequency representation that
preserves the temporal localization of features. CWT-based scalograms
have been used for PPG signal transformation to classify BP
(Normal, Prehypertension, Stage 1 hypertension, and Stage 2 hypertension)
\cite{ref26,wu2021improving}, estimate heart rate variability (HRV) and signal quality \cite{ref28} as well as to detect AF \cite{cheng2020atrial}. \added{CWT’s scale-adaptive time–frequency resolution suits PPG’s non-stationary, multi-scale nature, and its rich scalograms representation consistently delivers strong performance across various tasks.}

\added{The STFT-based spectrogram representation has been used in several PPG-based predictions using DL \cite{Chen2021STFTPPG, Chen2021STFTPPG}. In this approach, the PPG signals were transformed into fixed-resolution time–frequency spectrograms using the STFT and processed using a ResNet architecture. This approach, Spectrogram-ResNet, makes it possible to  use the image-based convolutional neural networks by using a simpler time–frequency representation compared to the CWT process \cite{Chen2021STFTPPG}. The standard STFT approach was also used in \cite{Chen2021STFTPPG} for signal quality estimation.}

Several alternatives have been explored for image-based PPG representations \cite{cheng2020atrial, wu2021improving}. These include \deleted{the standard STFT approach used in \cite{Chen2021STFTPPG} for signal quality estimation,} direct conversion of the 1D PPG waveform into a structured 2D representation (e.g. multi-channel raw-PPG derivatives obtained by stacking padded PPG cycles (or beats)) which has also been applied for BP estimation \cite{ref63}, Gramian Angular Fields (GASF/GADF) \cite{ref64} to encode temporal correlations, Visibility Graphs \cite{ref65} providing a topological view of waveform structure, and Symmetric Projection Attractor Reconstruction (SPAR) attractor images which encapsulate the morphology and variability of the signal \cite{SPAR1,SPAR2}. The optimal choice depends on the downstream task, as each method has distinct strengths and limitations. \deleted{CWT’s scale-adaptive time–frequency resolution suits PPG’s non-stationary, multi-scale nature, and its rich scalograms representation consistently delivers strong performance across various tasks.}

\hypertarget{materials-and-methods}{%
\section{Materials and Methods}\label{materials-and-methods}}

\hypertarget{datasets}{%
\subsection{Datasets}\label{datasets}}

This study utilized the PPG data contained in the VitalDB dataset \cite{ref29} for
BP estimation and in the DeepBeat dataset \cite{ref30} for the AF detection task.

\textbf{VitalDB dataset.} The VitalDB dataset includes ECG, PPG, and invasive arterial blood pressure (ABP) signals collected using patient monitors from surgical patients \cite{ref31}. Wang et al. \cite{ref29} published the pre-processed
VitalDB dataset as a subset of the PulseDB dataset, from which we use the PPG signals of 10s length with a sampling frequency of 125 Hz, and reference systolic BP (SBP) and diastolic BP (DBP) values derived from the ABP signals. \added{We used the VitalDB subset instead of the entire PulseDB dataset as it reduced the dataset size and was shown to lead to good generalization to out-of-distribution data \cite{ref59}.}

\added{The dataset supports both calibration-based and calibration-free evaluation
protocols for assessing BP model generalizability, where calibration denotes
subject-specific adaptation using reference BP measurements without retraining
or fine-tuning the model, through samples from the same subject in the training and test sets following the PulseDB convention \cite{ref29}. This
differs from the calibration definitions used in some prior cuffless BP studies
(e.g., \cite{refJoung}), where calibration may involve explicit model adaptation
or fine-tuning. The calibration-free scenario ensures no patient overlap between training and test set and therefore assesses the generalization to unseen patients. }

We refer to the first scenario as \textbf{VitalDB `Calib'} and to the
second scenario as \textbf{VitalDB `CalibFree'}. To ensure comparability with literature results, we keep the original test sets intact but split
the original training sets into training, validation, and calibration
sets, where the latter is not considered in this study, mimicking the
way the respective test sets were created, i.e., defining validation
sets with/without patient overlap for VitalDB `Calib'/'CalibFree'. In
Table \ref{table_1}, we summarize the two considered subsets, where one sample
corresponds to a segment of 10s length.

\begin{longtable}[]{@{}
  >{\raggedright\arraybackslash}p{0.35\columnwidth}
  >{\centering\arraybackslash}p{0.28\columnwidth}
  >{\centering\arraybackslash}p{0.28\columnwidth}@{}}

\toprule
Subset & VitalDB `Calib' & VitalDB `CalibFree' \\
\midrule
\endfirsthead
\toprule
Subset & VitalDB `Calib' & VitalDB `CalibFree' \\
\midrule
\endhead
\bottomrule
\endlastfoot
Train (samples / subjects) & 418986 / 1293 & 416880 / 1158 \\
Validation (samples / subjects) & 40673 / 1293 & 32400 / 90 \\
Test (samples / subjects) & 51720 / 1293 & 57600 / 144 \\
Age (years, mean ± SD) & 58.9 ± 15.0 & 58.8 ± 15.0 \\
Sex (M\%) & 57.6 & 57.9 \\
SBP (mmHg, mean ± SD) & 115.4 ± 18.9 & 115.4 ± 18.9 \\
DBP (mmHg, mean ± SD) & 62.9 ± 12.0 & 62.9 ± 12.0 \\
\caption{Characteristics of the VitalDB subsets used for BP estimation.} \label{table_1}
\end{longtable}

\textbf{DeepBeat dataset.} For the AF classification task, we leverage the DeepBeat
dataset \cite{ref30}. The dataset comprises more than 500,000 25-second, 32 Hz
PPG segments, henceforth referred to as samples, from 175 individuals
(108 with AF, 67 without). PPG signals were collected using a
wrist-based PPG wearable device from cohorts of participants before
cardioversion, patients undergoing an exercise stress test, and during
daily life \cite{ref30}. In the original publication \cite{ref30}, due to an uneven
AF/non-AF distribution across splits, performance metrics were
overestimated, and the strong test scores did not reflect equivalent
success on the validation and training sets. To address these issues, we
implement a new data split. We ensure no overlap between sets by
redistributing subjects, thereby eliminating the redundancy present in
the original dataset (Table \ref{table_2}). This revised split maintains an equal AF/non-AF
ratio across the training, validation, and test sets, thus providing a more
reliable representation and enhancing the robustness of our model
evaluation. (For more details,
please refer to the Supplementary Material.)

\begin{longtable}[]{@{}
  >{\raggedright\arraybackslash}p{0.2\columnwidth}
  >{\centering\arraybackslash}p{0.2\columnwidth}
  >{\centering\arraybackslash}p{0.15\columnwidth}
  >{\centering\arraybackslash}p{0.12\columnwidth}
  >{\centering\arraybackslash}p{0.12\columnwidth}@{}}

\toprule
Dataset & \multicolumn{4}{c}{DeepBeat (AF classification)} \\
\midrule
Subset & AF & Non-AF & Data Ratio & AF Ratio \\
\midrule
\endhead
\bottomrule
\endlastfoot
Train (samples / subjects) & 40603 / 50 & 65646 / 38 & 0.78 & 0.38 \\
Validation (samples / subjects) & 5800 / 19 & 9456 / 7 & 0.11 & 0.38 \\
Test (samples / subjects) & 5797 / 19 & 9580 / 5 & 0.11 & 0.37 \\

\caption{Characteristics of the DeepBeat subsets used for AF classification. ``Data Ratio” indicates the proportion of the full dataset assigned to each subset, and “AF Ratio” denotes the fraction of samples within that subset labeled as AF.} \label{table_2} \\ \\
\end{longtable}

\hypertarget{performance-evaluation-and-metrics}{%
\subsection{Performance Evaluation and
Metrics}\label{performance-evaluation-and-metrics}}

To keep the main text concise, we summarize our evaluation methods here.
For a detailed explanation of the procedures and metrics used, please
refer to the Supplementary Material.

\textbf{BP Estimation.} We evaluate BP predictions using the
Mean Absolute Error (MAE) and Mean Absolute Scaled Error (MASE),
comparing model outputs against a baseline that predicts the training
set median. Metrics are reported separately for systolic and diastolic
pressures.

\added{Furthermore, to determine the agreement between predictions and reference values in more detail, Bland-Altman analyses were also performed \cite{bland1986statistical}. The bias was determined as the mean difference between predictions and reference values, as well as the corresponding limits of agreement (LoA). The standard deviation of these differences was multiplied by 1.96 to determine the LoA. While MAE roughly indicates the size of errors, bias directly indicates systematic errors, while the LoA gives insight into the variability of individual prediction errors.}

We also provide the grading (A, B, C, D) based on IEEE 1708a-2019
standard \cite{ref32} which are calculated based on the difference between the
device or model's BP predictions and the reference (cuff-based)
measurements;

\begin{itemize}
\item
  \textbf{Grade A}: Errors $\leq$ 5 mmHg
\item
  \textbf{Grade B}: Errors $>$ 5 mmHg and $\leq$ 6 mmHg
\item
  \textbf{Grade C}: Errors $>$ 6 mmHg and $\leq$ 7 mmHg
\item
  \textbf{Grade D}: Errors \textgreater{} 7 mmHg
\end{itemize}

\added{We acknowledge that there are other BP estimation standards such as AAMI \cite{stergiou2018universal} and BHS \cite{o1993british}. In this manuscript, however, we restrict ourselves to the IEEE standard and defer a detailed comparison according to different BP estimation standards to future work. }

\textbf{AF Detection.} AF detection performance is
assessed using standard metrics: sensitivity, specificity, receiver
operator characteristic (ROC) area under the curve (AUC), F1 score, and
Matthews correlation coefficient (MCC). Classification thresholds are
adjusted to meet desired sensitivity/specificity criteria, with a
default threshold of 0.5 for the F1 score and optimized thresholds for
other metrics. We explore two different threshold choices by fixing the
threshold such that either sensitivity or specificity exceeds 0.8.

\added{\textbf{Statistical significance.} For every experimental setting, the best model, defined as the model with the lowest MAE for the regression tasks or the highest AUC for the classification tasks, was taken as the reference model. We then computed the performance difference between the reference model and every other model by performing 1,000 bootstraps. A performance difference whose 95\% confidence interval did not contain zero indicated a model to be statistically significantly worse. Conversely, if zero was in the interval, then the method indicated that the respective model was not statistically significantly worse than the reference, which was indicated in the result tables accordingly.}

\hypertarget{prediction-models}{%
\subsection{Prediction models}\label{prediction-models}}

In our study, we explore the use of three different input
representations -- time series, feature-based, and image-based
approaches -- each providing unique advantages in capturing the
underlying physiological information. We consider several architectures,
each designed to process either raw signals, extracted features, or
visual representations to accurately estimate BP values and classify
heart rhythms. All models were trained from scratch. 
\replaced{ The learning rate, as the most important hyperparameter, was systematically optimized based on the validation set performance, while model‑dependent hyperparameters were deliberately not extensively tuned to preserve a balanced experimental setting.}{No excessive hyperparameter tuning beyond exploratory investigations based on validation set performance was performed in any of the cases.} Due to the large number of considered model architectures, a more systematic exploration of hyperparameter choices was considered infeasible given the computational resources available for this study. Retrospectively, we see rather comparable performance levels across different input representations and involving independent code bases as an indication that the models/hyperparameter choices were not suboptimal.

\textbf{Baseline Models.} On VitalDB CalibFree, the baseline model (used
for BP estimation) predicts BP by outputting the median SBP/DBP value of
the BP data inferred from the training set for any given
input, providing a straightforward reference for evaluating the
performance of more advanced predictive models. On VitalDB Calib, we use
The subject-specific median was calculated as the prediction on the test set.

\textbf{Raw Time Series Models.} For both BP estimation and AF
detection, deep learning architectures such as CNNs, RNNs, and temporal convolutional networks (TCNs) have
been used to process raw ECG or PPG sequences to capture complex
temporal patterns, directly predicting continuous values for regression
or output probabilities for classification. The specific models used in
this study were:

\begin{itemize}
\item
  \textbf{LeNet1d:} A one-dimensional CNN adapted from the original
  LeNet architecture for feature extraction from raw ECG/PPG time-series
  data \cite{wagner2024explaining}.
\end{itemize}

\begin{itemize}
\item
  \textbf{Inception1d:} A 1D adaptation of the Inception architecture
  that uses parallel convolutional layers to capture multi-scale
  temporal features \cite{ismail2020inceptiontime}.
\item
  \textbf{XResNet1d50/101:} Deep residual networks modified for 1D data,
  incorporating group normalization and selective kernel sizes to learn
  hierarchical features from physiological signals \cite{ref35}.
\item
  \textbf{XResNet1d50+GNNLL:} This case uses the XResNet50d model with a
  Gaussian Negative Log Likelihood Loss (GNLL) instead of the
  conventional MAE loss as proposed in \cite{ref36}.
\item
  \textbf{AlexNet1d:} A one-dimensional variant of AlexNet that
  processes time series data through convolutional and pooling layers
  for regression and classification tasks \cite{ref38}.
\item
  \textbf{MiniRocket:} A nearly deterministic transform using dilated
  convolutions and a linear classifier, offering fast and effective
  feature extraction from time-series data \cite{ref39}.
\item
  \textbf{Temporal Convolutional Networks (TCNs):} Networks using
  dilated causal convolutions and residual connections to capture
  long-range dependencies in sequential physiological signals \cite{ref40}.

\item
    \textbf{PPNet:} A hybrid architecture consisting of CNN and LSTM layers
      designed to extract spatial and temporal representations from
      physiological time series \cite{panwar2020pp}.
\item
  \textbf{iTransformer:} A transformer-based model tailored for time
  series prediction and analysis for instance-specific
  normalization techniques and attention mechanisms for improved sequence modelling \cite{liu2023itransformer}.
\item
  \textbf{TimesNet:} A convolutional inception-like architecture for
  time series analysis which uses multi-periodic modelling by
  convolutional blocks for extracting detailed temporal patterns
  \cite{wu2022timesnet}.
\end{itemize}

We can roughly categorize the considered models into complex/deep models
(Inception1D, XResNet1d50, XResNet1d101, AlexNet1d, TCNs ) and simple/shallow models
(LeNet1d, MiniRocket).

\textbf{Feature-Based Models.} Clinically interpretable features are
extracted from PPG signals, such as pulse morphology metrics for BP
estimation and irregularity measures for AF detection, and fed into
machine learning algorithms to perform regression or classification
tasks with clearer interpretability. The specific models and techniques
used in this study were:

\begin{itemize}
\item
  \textbf{Clinically interpretable features (CIF):} CIF for BP includes
  features such as systolic peaks, diastolic peaks, pulse duration, and
  pulse morphology metrics \cite{ref41}. These features reflect vascular health,
  haemodynamic dynamics, and arterial compliance, making them
  well-suited for modelling BP. CIF for AF comprises features related to
  rhythm irregularity, such as randomness \cite{ref42}, variability \cite{ref43,ref44}, and
  complexity in inter-beat-intervals \cite{ref45}. These features highlight the
  irregular heart rhythms characteristic of AF. \added{Table \ref{tab:feature_categories} provides an overview of the hand-crafted features used in the
feature-based models.}The
  full list of features is described in the supplementary material.
  These features are combined with the following prediction models:

\begin{itemize}
\item
  \textbf{Multi-layer perceptron (MLP):} An MLP is a fully connected
  feedforward neural network that maps input features to target outputs
  through multiple layers of neurons.
\item
  \textbf{Gaussian Process Regression (GPR):} A non-parametric
  regression method that models complex relationships between PPG
  features and BP, providing probabilistic predictions \cite{ref46}.
\end{itemize}

\item
  \textbf{Wavelet Transformation:} A Wavelet-based method such as
  Wavelet Packet Decomposition using the Discrete Wavelet Transform \cite{ref47}
  is applied to the raw PPG signal. This transformation decomposes the
  signal into time-frequency components, extracting features that
  capture both transient and long-term patterns in the data. It should
  be noted that Wavelet + MLP forms a pipeline
  where the wavelet transform serves as a sophisticated feature
  extractor, and the MLP functions as the predictive model utilizing
  those features.
\end{itemize}

It should be noted that only the CIF directly reflects physiological mechanisms, e.g.\ pulse morphology or rhythm irregularity. On the other hand, features derived from mathematical transformations (e.g., \ wavelet coefficients) provide useful signal representations; however, they are not directly interpretable in clinical terms.

% \begin{table}[t]
% \centering
% \caption{\added{Overview of feature categories used in the feature-based models.}}
% \label{tab:feature_based_features}
% \begin{tabular}{ll}
% \toprule
% \textbf{Category} & \textbf{Extracted Features} \\
% \midrule
% Time-domain morphology & Pulse amplitude, rise time, decay time, area under pulse \\
% Temporal variability & Inter-beat intervals, heart rate, HR variability metrics \\
% Derivative-based & First and second derivative extrema and timings \\
% Frequency-domain & Spectral power in predefined frequency bands \\
% Quality-related & Signal quality indices and beat consistency measures \\
% \bottomrule
% \end{tabular}
% \end{table}

\begin{table}[t]
\centering
\caption{\added{Overview of feature categories and corresponding feature-based pipelines used for BP estimation and AF detection.}}
\label{tab:feature_categories}
\renewcommand{\arraystretch}{1.3}
\begin{tabular}{llccp{6.0cm}}
\hline
\textbf{Pipeline} & \textbf{Feature category} & \textbf{BP} & \textbf{AF} & \textbf{Description / Examples} \\
\hline
CIF 
& Time-domain morphology 
& \checkmark & -- 
& Pulse amplitude, rise time, decay time, pulse duration, area under the pulse \\

CIF 
& Derivative-based 
& \checkmark & -- 
& First and second derivative extrema, slope-based indices \\

CIF 
& Temporal variability 
& -- & \checkmark 
& Inter-beat intervals (PP intervals), HR, HRV, rhythm irregularity measures \\

CIF 
& Frequency-domain 
& -- & \checkmark 
& Spectral power in predefined frequency bands capturing rhythm variability \\

CIF 
& Quality-related 
& \checkmark & \checkmark 
& Signal quality indices, beat consistency measures \\

Wavelet + MLP 
& Wavelet-based features 
& \checkmark & \checkmark 
& Wavelet packet decomposition coefficients used as task-agnostic feature representations \\
\hline
\end{tabular}
\end{table}

%\begin{itemize}
%\item
%  \textbf{Image-Based Models: Continuous Wavelet Transform (CWT)
%  Scalograms:} PPG signals are transformed into time-frequency imagesFhyper
%  derived from PPG signals that capture localized signal variations
%  (26). The CWT-based images are used as inputs for ResNet18 models.
%\end{itemize}

\textbf{Image-Based Models}: One-dimensional signals can be converted into two-dimensional images, which capture the essence of the signal in a compact domain. Traditional image recognition CNNs can then be used with these image inputs. There are many different ways of converting signals to images, but this study only used the following two approaches:
\begin{itemize}
\item
\textbf{Continuous Wavelet Transform (CWT)
Scalograms:} PPG signals are transformed into time-frequency images
derived from PPG signals that capture localized signal variations \cite{ref26}. The CWT-based images are used as inputs for ResNet18 models \cite{he2016deep}.
\item
\added{\textbf{STFT-Based Spectrograms (Spectrogram-ResNet):}
PPG signals are transformed into fixed-resolution time-frequency spectrograms using the STFT \cite{Chen2021STFTPPG}. The resulting spectrograms are processed using convolutional neural network architectures based on ResNet-50 and ResNet-18, initialized with ImageNet pre-trained weights \cite{Deng2009ImageNet,He2016ResNet}.}

\end{itemize}

Table \ref{table_param_counts} lists overall parameter counts of all models within the tasks of AF detection and BP estimation, and is condensed by input representation (T/F/I). \added{It is worth noting that parameter count is reported as a coarse, hardware-agnostic
indicator of model complexity and is used in this work as a relative measure of
computational efficiency, but should not be interpreted as a direct proxy for deployment
cost on wearable or embedded hardware, where inference latency, memory footprint, and
energy consumption are strongly platform dependent.} For more details on these models and their implementations, please see
the Supplementary Material.

\hypertarget{results}{%
\section{\texorpdfstring{Results }{Results }}\label{results}}

\hypertarget{blood-pressure-estimation}{%
\subsection{BP Estimation}\label{blood-pressure-estimation}}

Tables \ref{table_3a},  \ref{table_3b},  \ref{table_4a}, and \ref{table_4b} display the results of BP (SBP/DBP) prediction
using different deep learning models based on VitalDB Calib and VitalDB
CalibFree datasets, respectively. The baseline models achieved MAEs of
14.87 and 9.43 mmHg for SBP and DBP, respectively, on CalibFree, and 10.72
and 5.78 mmHg, respectively, on Calib. On Calib, the per-subject baseline
performed substantially better than the global baseline (10.72 and 5.78
mmHg vs. 14.91 and 9.52 mmHg respectively).

Model performance varied between models and tasks. On CalibFree, almost
all models provided an improvement over baseline for both SBP and DBP
(the only exception being the Inception1d SBP model). However, the level
of improvement was moderate at best, with the lowest MASEs of 0.83 for
both SBP and DBP achieved by XResNet1d50, indicating 17\% reductions in
MAE in comparison to baseline. This resulted in at best 26\% of SBP
estimates and 40\% of DBP estimates falling into the top grade
(i.e.\ errors $\leq$ 5 mmHg). On Calib, there was greater
variation in model performances, with XResNet1d50+GNLL achieving the
lowest MASEs of 0.73 and 0.87 for SBP and DBP, respectively,
corresponding to 48\% of SBP estimates and 64\% of DBP estimates falling
into the top grade. In contrast, several models performed worse than the
per-subject baseline, with the worst-performing model, CIF+MLP,
achieving MASEs of 1.27 and 1.54 for SBP and DBP, respectively. Indeed,
for DBP estimation, only the XResNet1d50+GNLL model achieved an
improvement over the subject-specific baseline, whereas all others
performed worse than this baseline. Absolute performance in terms of MAE
was better on Calib than CalibFree, as shown by lower MAEs on Calib than
CalibFree for all models except the TCN+MLP SBP and DBP models, and the
CIF+MLP DBP model. MLP models always performed worse than other models
of the same type.

\begin{longtable}[]{@{}
  >{\raggedright\arraybackslash}p{0.02\columnwidth}
  >{\raggedright\arraybackslash}p{0.22\columnwidth}
  >{\centering\arraybackslash}p{0.12\columnwidth}
  >{\centering\arraybackslash}p{0.10\columnwidth}
  >{\centering\arraybackslash}p{0.10\columnwidth}
  >{\centering\arraybackslash}p{0.08\columnwidth}
  >{\centering\arraybackslash}p{0.08\columnwidth}
  >{\centering\arraybackslash}p{0.08\columnwidth}
  >{\centering\arraybackslash}p{0.08\columnwidth}@{}}

\toprule
\multirow{2}{*}{} & \multirow{2}{*}{Model}
& \multicolumn{3}{c}{SBP}
& \multicolumn{4}{c}{IEEE Grades (SBP)} \\
\cmidrule(lr){3-5} \cmidrule(lr){6-9}
& & MAE (MASE) & Bias & LoA
& A & B & C & D \\
\midrule
\endhead
\bottomrule
\endlastfoot

\multirow{2}{*}{\textbf{B}}
& Baseline (global)
& 14.91 (1.39) & 0.00 & 36.94 & 0.21 & 0.04 & 0.04 & 0.71 \\

& Baseline (per subject)
& 10.72 (1.00) & 0.00 & 28.01 & 0.34 & 0.06 & 0.05 & 0.55 \\
\midrule

\multirow{9}{*}{\textbf{T}}
& XResNet1d101
& 9.06 (0.84) & 0.25 & 24.51 & 0.40 & 0.06 & 0.06 & 0.48 \\

& XResNet1d50
& 9.52 (0.88) & 0.29 & 25.00 & 0.37 & 0.06 & 0.06 & 0.51 \\

& Inception1d
& 9.43 (0.88) & -0.90 & 24.55 & 0.36 & 0.06 & 0.06 & 0.52 \\

& LeNet1d
& 11.76 (1.09) & -0.78 & 29.40 & 0.27 & 0.05 & 0.05 & 0.63 \\

& XResNet1d50 +GNLL
& \textbf{\ul{7.99 (0.74)}} & -0.79 & 22.44 & 0.48 & 0.06 & 0.05 & 0.41 \\

& AlexNet1d
& 9.49 (0.88) & -0.54 & 25.21 & 0.38 & 0.06 & 0.06 & 0.50 \\

& Minirocket
& 11.24 (1.05) & 0.11 & 28.39 & 0.29 & 0.06 & 0.05 & 0.60 \\

& iTransformer
& 12.66 (1.18) & -0.35 & 31.71 & 0.25 & 0.05 & 0.05 & 0.65 \\

& TimesNet
& 12.69 (1.16) & -0.71 & 31.88 & 0.25 & 0.05 & 0.05 & 0.65 \\

& PPNet
& 8.76 (0.81) & -0.42 & 23.63 & 0.41 & 0.06 & 0.06 & 0.47 \\

& TCN +MLP
& 12.84 (1.19) & -5.50 & 30.71 & 0.25 & 0.05 & 0.05 & 0.65 \\

\midrule

\multirow{3}{*}{\textbf{F}}
& WAVELET +MLP
& 13.62 (1.26) & -1.77 & 34.10 & 0.24 & 0.05 & 0.04 & 0.67 \\

& CIF +GPR
& 12.22 (1.13) & -0.32 & 32.03 & 0.27 & 0.05 & 0.05 & 0.63 \\

& CIF +MLP
& 13.78 (1.27) & -0.13 & 34.21 & 0.23 & 0.04 & 0.04 & 0.69 \\
\midrule

\multirow{2}{*}{\textbf{I}}
& CWT
& 10.91 (1.01) & -0.24 & 28.51 & 0.32 & 0.06 & 0.05 & 0.57 \\

& Spectrogram-ResNet-18
& 9.85 (0.91) & -1.14 & 26.83 & 0.38 & 0.06 & 0.06 & 0.50 \\

& Spectrogram-ResNet-50
& 9.82 (0.91) & -0.81 & 26.88 & 0.39 & 0.06 & 0.06 & 0.49 \\

\caption{The SBP performance on the VitalDB Calib dataset for three input representations, T for raw time series, F for feature-based, and I for image-based models, next to B for the baseline model.
\added{The best-performing model is marked in bold and underlined. Models that do not perform statistically significantly worse than the best model are shown in bold. All MAE values are given in units of mmHg. Bias and LoA were computed following Bland--Altman analysis (bias $\pm$ 1.96 SD).}}
\label{table_3a}
\end{longtable}

\begin{longtable}[]{@{}
  >{\raggedright\arraybackslash}p{0.02\columnwidth}
  >{\raggedright\arraybackslash}p{0.22\columnwidth}
  >{\centering\arraybackslash}p{0.12\columnwidth}
  >{\centering\arraybackslash}p{0.10\columnwidth}
  >{\centering\arraybackslash}p{0.10\columnwidth}
  >{\centering\arraybackslash}p{0.08\columnwidth}
  >{\centering\arraybackslash}p{0.08\columnwidth}
  >{\centering\arraybackslash}p{0.08\columnwidth}
  >{\centering\arraybackslash}p{0.08\columnwidth}@{}}

\toprule
\multirow{2}{*}{} & \multirow{2}{*}{Model}
& \multicolumn{3}{c}{DBP}
& \multicolumn{4}{c}{IEEE Grades (DBP)} \\
\cmidrule(lr){3-5} \cmidrule(lr){6-9}
& & MAE (MASE) & Bias & LoA
& A & B & C & D \\
\midrule
\endhead
\bottomrule
\endlastfoot

\multirow{2}{*}{\textbf{B}}
& Baseline (global)
& 9.52 (1.65) & 0.00 & 23.66 & 0.32 & 0.06 & 0.06 & 0.56 \\

& Baseline (per subject)
& 5.78 (1.00) & 0.00 & 15.37 & 0.56 & 0.07 & 0.06 & 0.30 \\
\midrule

\multirow{9}{*}{\textbf{T}}
& XResNet1d101
& 5.91 (1.02) & -0.06 & 15.86 & 0.55 & 0.07 & 0.06 & 0.32 \\

& XResNet1d50
& 6.33 (1.08) & 0.02 & 16.51 & 0.50 & 0.08 & 0.07 & 0.35 \\

& Inception1d
& 6.37 (1.11) & -0.02 & 16.46 & 0.50 & 0.07 & 0.07 & 0.36 \\

& LeNet1d
& 7.80 (1.35) & 1.00 & 19.34 & 0.39 & 0.07 & 0.07 & 0.47 \\

& XResNet1d50 +GNLL
& \textbf{\ul{5.09 (0.87)}} & -0.52 & 14.56 & 0.64 & 0.06 & 0.05 & 0.25 \\

& AlexNet1d
& 6.10 (1.05) & -0.14 & 16.33 & 0.54 & 0.07 & 0.06 & 0.33 \\

& Minirocket
& 7.33 (1.26) & 0.01 & 18.51 & 0.43 & 0.07 & 0.07 & 0.43 \\

& iTransformer
& 8.20 (1.41) & -0.18 & 20.53 & 0.38 & 0.07 & 0.06 & 0.49 \\

& TimesNet
& 8.12 (1.40) & 0.01 & 20.69 & 0.38 & 0.07 & 0.06 & 0.49 \\

& PPNet
& 5.58 (1.01) & -0.18 & 15.23 & 0.58 & 0.07 & 0.06 & 0.29\\ 

& TCN +MLP
& 8.48 (1.46) & -3.12 & 20.46 & 0.37 & 0.07 & 0.07 & 0.49 \\

\midrule

\multirow{3}{*}{\textbf{F}}
& WAVELET +MLP
& 8.84 (1.51) & -1.00 & 22.08 & 0.36 & 0.07 & 0.06 & 0.51 \\

& CIF +GPR
& 7.78 (1.35) & -0.10 & 19.69 & 0.40 & 0.07 & 0.07 & 0.46 \\

& CIF +MLP
& 8.94 (1.54) & -0.55 & 22.18 & 0.35 & 0.06 & 0.06 & 0.53 \\
\midrule

\multirow{2}{*}{\textbf{I}}
& CWT
& 6.68 (1.15) & -0.32 & 17.66 & 0.50 & 0.07 & 0.06 & 0.37 \\

& Spectrogram-ResNet-18
& 6.33 (1.09) & -0.64 & 17.21 & 0.54 & 0.07 & 0.06 & 0.33 \\

& Spectrogram-ResNet-50
& 6.39 (1.10) & -0.10 & 17.42 & 0.53 & 0.07 & 0.06 & 0.34 \\

\caption{The DBP performance on the VitalDB Calib dataset for three input representations, T for raw time series, F for feature-based, and I for image-based models, next to B for the baseline model.
\added{The best-performing model is marked in bold and underlined. Models that do not perform statistically significantly worse than the best model are shown in bold. All MAE values are given in units of mmHg. Bias and LoA were computed following Bland--Altman analysis (bias $\pm$ 1.96 SD).}}
\label{table_3b}
\end{longtable}

\begin{longtable}[]{@{}
  >{\raggedright\arraybackslash}p{0.02\columnwidth}
  >{\raggedright\arraybackslash}p{0.24\columnwidth}
  >{\centering\arraybackslash}p{0.12\columnwidth}
  >{\centering\arraybackslash}p{0.07\columnwidth}
  >{\centering\arraybackslash}p{0.07\columnwidth}
  >{\centering\arraybackslash}p{0.05\columnwidth}
  >{\centering\arraybackslash}p{0.05\columnwidth}
  >{\centering\arraybackslash}p{0.05\columnwidth}
  >{\centering\arraybackslash}p{0.05\columnwidth}@{}}

\toprule
\multirow{2}{*}{} & \multirow{2}{*}{Model}
& \multicolumn{3}{c}{SBP}
& \multicolumn{4}{c}{IEEE Grades (SBP)} \\
\cmidrule(lr){3-5} \cmidrule(lr){6-9}
& & MAE (MASE) & Bias & LoA
& A & B & C & D \\
\midrule
\endhead
\bottomrule
\endlastfoot

\textbf{B}
& Baseline
& 14.87 (1.00) & -0.00 & 36.94 & 0.21 & 0.04 & 0.04 & 0.71 \\
\midrule

\multirow{12}{*}{\textbf{T}}
& XResNet1d101
& 12.43 (0.83) & -1.28 & 30.91 & 0.25 & 0.05 & 0.05 & 0.65 \\

& XResNet1d50
& 12.46 (0.83) & -0.23 & 31.09 & 0.24 & 0.05 & 0.05 & 0.66 \\

& Inception1d
& 14.46 (0.97) & -3.15 & 30.62 & 0.25 & 0.05 & 0.05 & 0.65 \\

& LeNet1d
& 12.37 (0.83) & 0.86 & 30.57 & 0.25 & 0.05 & 0.05 & 0.65 \\

& XResNet1d50+GNLL
& \textbf{\ul{12.27 (0.82)}} & -0.99 & 30.55 &0.26 & 0.05 & 0.05 & 0.64 \\

& AlexNet1d
& 12.51 (0.84) & 0.52 & 31.24 & 0.25 & 0.05 & 0.05 & 0.65 \\

& Minirocket
& 12.65 (0.85) & -0.95 & 31.43 & 0.26 & 0.05 & 0.05 & 0.64 \\

& iTransformer
& 13.15 (0.88) & -0.75 & 32.76 & 0.24 & 0.05 & 0.05 & 0.66 \\

& TimesNet
& 13.09 (0.88) & -1.89 & 32.84 & 0.25 & 0.05 & 0.05 & 0.65 \\

& PPNet
& 12.56 (0.84) & -2.05 & 31.24 & 0.25 & 0.05 & 0.05 & 0.65 \\

& TCN +MLP
& 12.72 (0.85) & -2.52 & 31.77 & 0.25 & 0.05 & 0.05 & 0.65 \\
\midrule

\multirow{3}{*}{\textbf{F}}
& WAVELET +MLP
& 14.21 (0.95) & 0.03 & 35.32 & 0.22 & 0.04 & 0.04 & 0.70 \\

& CIF +GPR
& 12.90 (0.87) & -0.32 & 32.38 & 0.25 & 0.05 & 0.04 & 0.66 \\

& CIF +MLP
& 14.02 (0.94) & 0.15 & 34.82 & 0.24 & 0.04 & 0.04 & 0.68 \\
\midrule

\multirow{2}{*}
& CWT
& 13.49 (0.90) & -1.14 & 33.63 & 0.23 & 0.05 & 0.04 & 0.68 \\

& Spectrogram-ResNet-18
& 13.01 (0.87) & 0.32 & 32.29 & 0.24 & 0.04 & 0.04 & 0.68 \\
{\textbf{I}}
& Spectrogram-ResNet-50
& 13.53 (0.90) & 3.40 & 32.88 & 0.24 & 0.04 & 0.04 & 0.68 \\

\caption{The SBP performance on the VitalDB CalibFree dataset for three input representations, T for raw time series, F for feature-based, and I for image-based models, next to B for the baseline model.
\added{The best-performing model is marked in bold and underlined. Models that do not perform statistically significantly worse than the best model are shown in bold. All MAE values are given in units of mmHg. Bias and LoA were computed following Bland--Altman analysis (bias $\pm$ 1.96 SD).}}
\label{table_4a}
\end{longtable}

\begin{longtable}[]{@{}
  >{\raggedright\arraybackslash}p{0.02\columnwidth}
  >{\raggedright\arraybackslash}p{0.24\columnwidth}
  >{\centering\arraybackslash}p{0.12\columnwidth}
  >{\centering\arraybackslash}p{0.07\columnwidth}
  >{\centering\arraybackslash}p{0.07\columnwidth}
  >{\centering\arraybackslash}p{0.05\columnwidth}
  >{\centering\arraybackslash}p{0.05\columnwidth}
  >{\centering\arraybackslash}p{0.05\columnwidth}
  >{\centering\arraybackslash}p{0.05\columnwidth}@{}}

\toprule
\multirow{2}{*}{} & \multirow{2}{*}{Model}
& \multicolumn{3}{c}{DBP}
& \multicolumn{4}{c}{IEEE Grades (DBP)} \\
\cmidrule(lr){3-5} \cmidrule(lr){6-9}
& & MAE (MASE) & Bias & LoA
& A & B & C & D \\
\midrule
\endhead
\bottomrule
\endlastfoot

\textbf{B}
& Baseline
& 9.43 (1.00) & 0.00 & 23.66 & 0.33 & 0.06 & 0.06 & 0.55 \\
\midrule

\multirow{12}{*}{\textbf{T}}
& XResNet1d101
& \textbf{7.93 (0.84)} & 0.18 & 19.70 & 0.39 & 0.07 & 0.06 & 0.48 \\

& XResNet1d50
& 7.98 (0.84) & -1.55 & 19.63 & 0.39 & 0.07 & 0.06 & 0.48 \\

& Inception1d
& \textbf{7.94 (0.84)} & -1.86 & 19.29 & 0.38 & 0.07 & 0.06 & 0.49 \\

& LeNet1d
& \ul{\textbf{7.92 (0.84)}} & 1.26 & 19.45 & 0.39 & 0.07 & 0.06 & 0.48 \\

& XResNet1d50+GNLL
& \textbf{7.94 (0.84)} & -0.54 & 19.63 & 0.39 & 0.07 & 0.06 & 0.48 \\

& AlexNet1d
& 7.98 (0.84) & 0.15 & 19.80 & 0.38 & 0.07 & 0.07 & 0.48 \\

& Minirocket
& 8.08 (0.85) & -1.01 & 19.87 &0.38 & 0.07 & 0.06 & 0.49 \\

& iTransformer
& 8.37 (0.88) & -0.67 & 20.77 & 0.37 & 0.07 & 0.06 & 0.50 \\

& TimesNet
& 8.29 (0.87) & -0.71 & 20.79 & 0.38 & 0.07 & 0.06 & 0.49 \\

& PPNet
& \textbf{7.95 (0.84)} & -0.91 & 19.74 & 0.39 & 0.07 & 0.07 & 0.48 \\

& TCN +MLP
& 8.24 (0.87) & -1.77 & 20.27 & 0.38 & 0.06 & 0.06 & 0.50 \\
\midrule

\multirow{3}{*}{\textbf{F}}
& WAVELET +MLP
& 8.89 (0.94) & -0.78 & 22.10 & 0.35 & 0.06 & 0.06 & 0.53 \\

& CIF +GPR
& 8.15 (0.86) & -0.44 & 20.41 & 0.38 & 0.07 & 0.06 & 0.49 \\

& CIF +MLP
& 8.69 (0.92) & 0.38 & 21.61 & 0.36 & 0.06 & 0.06 & 0.52 \\
\midrule

\multirow{2}{*}
& CWT
& 8.41 (0.89) & -0.18 & 20.96 & 0.37 & 0.07 & 0.07 & 0.49 \\
{\textbf{I}}
& Spectrogram-ResNet-18
& 8.04 (0.85) & 0.27 & 19.92 & 0.38 & 0.07 & 0.07 & 0.48 \\

& Spectrogram-ResNet-50
& 8.34 (0.88) & 2.00 & 20.30 & 0.37 & 0.07 & 0.06 & 0.50 \\

\caption{The DBP performance on the VitalDB CalibFree dataset for three input representations, T for raw time series, F for feature-based, and I for image-based models, next to B for the baseline model.
\added{The best-performing model is marked in bold and underlined. Models that do not perform statistically significantly worse than the best model are shown in bold. All MAE values are given in units of mmHg. Bias and LoA were computed following Bland--Altman analysis (bias $\pm$ 1.96 SD).}}
\label{table_4b}
\end{longtable}

Raw time series (T) and image-based (I) models performed better than
feature-based (F) models on Calib (with the one exception of TCN+MLP),
as shown by MASEs of 0.73-1.07 (SBP) and 0.87-1.46 (DBP) for raw time
series and image-based models, compared to 1.13-1.27 and 1.35-1.54 for
feature-based models. There was a less clear difference in performance
on CalibFree. Whilst best performance was achieved with raw time series
models, the image-based model performed better than three raw time
series models on Calib, and better than one raw time series model on
CalibFree. For Calib, more complex models seem to exhibit an advantage
(comparing, for example, XResNet1d50 and XResNet1d101), most likely due to
the ability to memorize subject-specific signal patterns. The only
exception from this pattern seems to be the AlexNet1d model, which shows
a performance that is almost on par with more complex models and
considerably better than comparable lightweight models (such as LeNet1d
or TCN+MLP).

Absolute performance in terms of MAE was always better for DBP
estimation than for SBP estimation. However, when considering errors
relative to baseline, MASEs were broadly similar between SBP and DBP on
CalibFree, and always higher for DBP than SBP on Calib. Since the IEEE
grading system uses absolute errors, performance in terms of IEEE Grades
was generally better for DBP than SBP.

\hypertarget{atrial-fibrillation-detection}{%
\subsection{AF
Detection}\label{atrial-fibrillation-detection}}
Table \ref{table_5} represents the performance analysis of the classification task
(AF/ non-AF) based on the Deepbeat dataset. \added{Specificity (sensitivity $\geq 0.8$) and sensitivity (specificity $\geq 0.8$) are reported at operating thresholds selected to ensure a minimum sensitivity or specificity of 0.8, respectively.}

Similar to the regression task, raw time series (T) and image-based (I) models performed better
than feature-based (F) models in all cases except for the LeNet1d raw
time series model, which performed poorly. The best-performing models
were Inception1d and XResNet1d50, which achieved AUCs of 0.87, and F1
scores of 0.72 and 0.69 respectively. As with the regression task,
whilst best performance was achieved with the more complex raw time
series models, the image-based model performed better than some raw time
series models. Whilst feature-based models generally performed worst,
wavelet-based features produced better performance than clinically
interpretable features.

\begin{table}[ht]
\centering
\begin{tabular}{@{}
  >{\raggedright\arraybackslash}p{0.02\textwidth}
  >{\raggedright\arraybackslash}p{0.15\textwidth}
  >{\centering\arraybackslash}p{0.1\textwidth}
  >{\centering\arraybackslash}p{0.1\textwidth}
  >{\centering\arraybackslash}p{0.15\textwidth}
  >{\centering\arraybackslash}p{0.15\textwidth}
  >{\centering\arraybackslash}p{0.1\textwidth}
  >{\centering\arraybackslash}p{0.1\textwidth}
  >{\centering\arraybackslash}p{0.12\textwidth}@{}}
\toprule
 & Model & AUC & F1 (0.5) & Specificity (sensitivity > 0.8) & Sensitivity (specificity > 0.8) & MCC (sensitivity > 0.8) & MCC (specificity > 0.8) \\
\midrule
\multirow{7}{*}{\textbf{T}} 
& XResNet1d101 & \textbf{\ul{0.85}} & 0.66 & 0.74 & 0.74 & 0.53 & 0.53 \\
& XResNet1d50 & \textbf{0.85} & 0.69 & 0.75 & 0.74 & 0.54 & 0.54 \\
& Inception1d & \textbf{0.85} & 0.69 & 0.74 & 0.73 & 0.52 & 0.52  \\
& LeNet1d & 0.74 & 0.55 & 0.55 & 0.48 & 0.34 & 0.30  \\

& AlexNet1d & 0.82 & 0.65 & 0.68 & 0.65 & 0.46 & 0.45 \\
& Minirocket & 0.82 & 0.66 & 0.69 & 0.67 & 0.47 & 0.47 \\

& iTransformer & 0.68 & 0.40 & 0.47 & 0.36 & 0.27 & 0.17  \\
& TimesNet & 0.79 & 0.58 & 0.61 & 0.57 & 0.40 & 0.38 \\
& PPNet & \textbf{0.85} & 0.69 & 0.76 & 0.75 & 0.54 & 0.54 \\

& TCN+MLP & \textbf{0.85} & 0.67 & 0.75 & 0.74 & 0.54 & 0.53  \\
\midrule
\multirow{2}{*}{\textbf{F}} 
& WAVELET + MLP & 0.76 & 0.60 & 0.58 & 0.52 & 0.37 & 0.33  \\
& CIF+MLP & 0.52 & 0.39 & 0.20 & 0.31 & -0.002 & 0.13  \\
\midrule
\multirow{1}{*}
& CWT & 0.82 & 0.69 & 0.72 & 0.69 & 0.50 & 0.49  \\
\textbf{I}
& Spectrogram-ResNet-18 & 0.82 & 0.68 & 0.68 & 0.61 & 0.47 & 0.42  \\
& Spectrogram-ResNet-50 & \textbf{0.85} & 0.69 & 0.72 & 0.69 & 0.50 & 0.49  \\
\bottomrule
\end{tabular}
\caption{The performance analysis for the classification task on the DeepBeat dataset for three input representations: \textbf{T} for raw time series, \textbf{F} for feature-based, and \textbf{I} for image-based models. \added{The best-performing model is marked in bold and underlined. Models that do not perform statistically significantly worse than the best model are shown in bold.}}
\label{table_5}
\end{table}

\hypertarget{discussion}{%
\section{Discussion}\label{discussion}}

\hypertarget{blood-pressure-estimation-1}{%
\subsection{BP
Estimation}\label{blood-pressure-estimation-1}}

\textbf{Relative Performance Comparison.} For both regression tasks, the best results were achieved by approaches based on raw time series.
Feature-based approaches were not competitive in the Calib scenario, but are to a certain degree competitive in the CalibFree scenario. Here, it
is important to note the differences between Calib and CalibFree:
CalibFree tests on unseen patients, whereas Calib purposely tests on
patients already seen during training, i.e, .\ models can profit from a
certain degree of memorization. Quite naturally, models achieved much
better scores in the Calib setting (even though test sets are not
entirely comparable). This also impacts the best-forming models in each
of the two settings (within the category of models operating on raw time
series). Large and complex models (such as XResNet1d and Inception1d)
performed best on Calib, as overfitting to specific patients is
desirable, whereas smaller models (such as LeNet1d) performed on par or
even better than more complex models in the CalibFree scenario, where
generalization to unseen patients is key. 
This observation aligns with the fact that the gap between raw time series and feature-based
approaches is larger in the case of Calib, which is quite natural as
this task largely profits from the complexity of the model (and all
feature-based approaches are less complex than typical models operating
on raw time series). \added{Beyond aggregate error metrics, Bland-Altman analysis showed a small degree of systematic bias but large LoA regardless of model, indicating the presence of considerable variability at the sample level even in the best-performing models. This indicates that a smaller MAE does not necessarily imply a reliable prediction at the sample level.}

The set of considered raw time series models also included transformer or hybrid CNN-LSTM models. PPNet, in spite of having a fairly modest parameter size ($\sim$125k), was competitive in both Calib and CalibFree cases, coming close to matching the variants of XResNet and emphasizing the potential of light-weight hybrid CNN–LSTM models. Conversely, TimesNet ($\sim$667k parameters) and iTransformer ($\sim$642k parameters) only modestly improved on the baseline and did not reach CNN-level accuracy. Along with the parameter study in Table \ref{table_param_counts}, these indicate that model size does not directly correlate with results: AlexNet ($>$40m parameters) and TCNS ($>$6m parameters) are among the largest models and yet under-performed, yet the significantly smaller PPNet was more effective.

\textbf{Comparison to Literature Results.} The number of published
results on PulseDB is very limited. To the best of our knowledge \cite{ref48} is
the only prior study that reported results on PulseDB, albeit on the
full dataset and not just the VitalDB subset, achieving a MAE of 10-11
mmHg for SBP after self-supervised pretraining. We refer to \cite{ref49} for a
recent comparative analysis of different models on different datasets,
where models achieved MAEs of between 11 and 19 mmHg for SBP, and
between 7 and 11 mmHg for DBP. In comparison, in this study, the lowest
MAEs were in the range of approximately 12–13 mmHg and 8 mmHg for SBP and DBP, respectively, on VitalDB
CalibFree. This suggests that the presented best-performing models reached
state-of-the-art performance.

\textbf{Impact of Loss Function and Dropout Ensembling.} The performance
of the XResNet1d50+GNLL in comparison to the XResNet1d50 trained with a
regular MAE loss highlights the benefits of incorporating dropout
ensembling at evaluation, and using a likelihood-based loss can have on
model performance. It suggests that potentially CNN results
could be improved by using a different loss function and ensemble-based
evaluation procedure. Interestingly, this only applied to the Calib and
not the CalibFree scenario. Other works have reported that modelling
uncertainties may improve predictive performance on models trained on
ECG data \cite{ref50}. This variant had two parallel output branches with
dropout layers (with a constant but low dropout rate, e.g.\ 4-5\%
depending on the task/parameter) dispersed throughout the architecture.
Each test input was evaluated 100 times with the dropout layers left
active. The predictions produced with the technique are the average of
the distribution of outputs produced for each input. Also, a separate
model was trained to predict each BP type
(systolic/diastolic).

\textbf{Overall Performance.} The best-performing model
(XResNet1d50+GNLL on Calib) achieved up to 48\% of SBP estimates, and
64\% of DBP estimates in the top IEEE category (Grade A), i.e.\ with
prediction errors $\leq$ 5 mmHg. However, even with this model, 41\% of
SBP estimates, and 25\% of DBP estimates fell into the lowest category
(Grade D). For clinical applications, reducing the fraction of grade D
is highly important, or at least identifying patient characteristics
that allow to predict whether an unseen patient will belong to grade A
or D. This task is referred to as out-of-model-scope detection \cite{ref51}, a
task which is closely related to out-of-distribution detection.
Reliable uncertainty estimation is a commonly used approach for the
latter and therefore also represents a promising direction for future
research. More generally, it might be insightful to assess whether
coarser prediction targets, i.e.\ framing BP prediction as a
classification target, could provide clinically meaningful insights, see
\cite{ref52} for an exploration.

\textbf{Limitations of the VitalDB Dataset.} VitalDB is an
intraoperative vital signs dataset. Key advantages are that it includes
non-cardiac patients, and it is labelled with detailed information on
surgical type, surgical approach, anaesthetic type, duration, and
device. The dataset covers a wide age range, from neonates to adults.
For this study, it was treated as a single dataset, but further
disaggregation into subgroups would be essential to identify clinically
meaningful features and assess their generalizability in wider
healthcare settings. Furthermore, the dataset is not representative of
the key application of PPG-based BP monitoring in wearables in the
general population, because data were acquired from subjects during
surgery rather than in daily life, using clinical pulse oximeters rather
than wearable devices. \added{It is also worth noting that since the recordings were obtained from anesthetized patients in the controlled surgical settings, the PPG signals contain minimal motion artifacts, and hence, it might result in optimistic performance estimates when compared to real-world wearable scenarios.}

\hypertarget{atrial-fibrillation-detection-1}{%
\subsection{AF
Detection}\label{atrial-fibrillation-detection-1}}

\textbf{Relative Performance Comparison.} Best performance on the
classification case was achieved by modern CNN architectures (XResNet1d,
Inception), which performed at least on par with, and in some cases
showed advantages over, simpler models (LeNet1d). Feature-based
approaches (based on wavelets) showed comparable performance to simple
CNNs but were clearly outperformed by large-scale CNNs operating on raw
time series input. Furthermore, it is worth noting that although PPNet's performance was very close to the top-performing CNNs, it still fell short, whereas both iTransformer and TimesNet were distinctly inferior, highlighting the relative effectiveness of the convolutional architectures on this task.
\added{Notably, the STFT-based Spectrogram-ResNet achieved performance comparable
to the top-performing raw time-series CNNs in terms of AUC, F1-score, and
thresholded sensitivity and specificity, indicating that image-based
representations can be competitive for AF detection. 
}
\added{Furthermore, bootstrap analyses showed that several top-performing CNN and image-based models were statistically indistinguishable in terms of classification accuracy, suggesting that increasing architectural complexity yields diminishing returns for this data.}

\textbf{Comparison to Literature Results.} The original publication
describing the DeepBeat dataset \cite{ref30} reported F1-scores of 0.54 for
AF-prediction without pretraining and 0.71 for a multi-task model that
jointly predicted AF and signal quality. It is important to stress that
these results were achieved on a different, imbalanced split of the
dataset and are therefore not directly comparable to those in the
present study. The SiamQuality CNN model \cite{ref48} achieved F1-scores of up
to 0.71 on DeepBeat, albeit after self-supervised pretraining. To enable
a direct comparison, we train an XResNet1d50 model on the original
split and achieved an F1-score of 0.69, which is clearly superior to the
model presented in the original DeepBeat publication. This demonstrates
that the presented models reached state-of-the-art performance in
comparison to literature benchmarks.

\textbf{Overall Performance.} The best-performing models showed a strong
performance in terms of absolute performance values, with F1-scores
comparable to or better than reported values in the literature (using
the original splits provided by DeepBeat), and sensitivities and
specificities of around 0.8. It is worth noting that even though certain
commercial AF detection algorithms claim to reach 0.98 sensitivity at
$>$ 0.99 specificity on internal validation datasets \cite{ref53}, they
typically rely on ECG measurements, which serve as the gold standard for
AF detection, and severely overestimate the model performance under
real-world conditions as the reported performance only applies to
certain heart range intervals \cite{ref54}. In practice, PPG-based AF detection
algorithms in consumer wearables are often designed to provide a high
positive predictive value by requiring multiple predictions of AF before
raising an alert \cite{ref55,ref56}. Further work would be required to investigate
whether a high positive predictive value could be achieved on real-world
data using the models presented in this study.

\textbf{CIF vs. Raw Time Series Models.} The difference in AF detection
performance between raw time series models and those based on clinically
interpretable features may be influenced by multiple factors, including
biases in \added{the} data acquisition process, artifacts, and even errors in the
labelling of recordings into different classes. Deep learning-based raw
time series models can use a broad range of information from the signal,
enhancing their ability to detect class-specific patterns. However, the drawback is that clinically irrelevant features, such as artifacts and
biases, may be incorporated into the training process. In contrast,
models that rely on predefined clinically interpretable features, such
as pulse irregularity in AF, have a more constrained feature space.
While this improves interpretability and robustness in some cases, it
may also make these models more vulnerable to mislabelled or noisy data.
Initial experiments indicate that incorporating signal quality
assessments next to interpretable features can enhance prediction
performance. Future work should investigate the classification
performance in terms of more fine-grained subgroups, in particular
stratified according to signal and labelling quality. Such analysis
could reveal if feature-based methods exhibit advantages over raw time
series approaches in high-quality subsets, where the underlying
assumptions of clinical validity are most clearly satisfied.

\textbf{Limitations of the DeepBeat Dataset.} While the DeepBeat dataset
represents one of the few publicly available PPG-based AF prediction dataset that is large enough for training deep learning models, it is
very sparsely documented, and no corresponding ECG signals are
available, which would allow one to retrospectively assess the
annotation quality. For DeepBeat, three independent cohorts were
examined: patients admitted for cardioversion before treatment
(classified as AF based on patient ID rather than data window), different participants undergoing exercise stress tests and a challenge dataset as the only subset containing both AF and non-AF samples. Furthermore, the
data were categorized into low-, medium-, and high-quality segments, all
of which were included in this study. However, there was a notable
imbalance, with a greater number of high-quality AF-labelled segments
compared to non-AF-labelled segments.

Even though the DeepBeat authors demonstrated the generalization of
models trained on DeepBeat to external datasets, due to the substantial
differences between these datasets, we cannot conclusively attribute the
observed changes solely to AF. The PPG feature variations used to
classify the datasets may have resulted from other fundamental
differences between them.

\hypertarget{general-insights-and-directions-for-future-research}{%
\subsection{General Insights and Directions for Future
Research}\label{general-insights-and-directions-for-future-research}}

%\added{Across both classification and regression tasks, our findings reveal some common patterns that might inform future work. First, models learned from raw waveforms end-to-end invariably beat feature-based methods, arguably because they extract fine-scale morphology and temporal detail without any need for handengineered features. Second, large and deep models shine in cases when patient overlap provides some memorization room (e.g., calibration-based BP estimation), while shallow models hold their own when out-of-patient generalization is needed (e.g., calibration-free BP estimation). Third, image-based methods like wavelet scalograms performed competitively in some cases, implying that they can be a complementary option when interpretability or time–frequency analysis is of importance. These findings give some general guidelines for selecting model families based on target task and dataset properties.}

\textbf{General Recommendations on Model Choices.} In this study,
\replaced{modern CNN architectures such as XResNet1d provided the best performance in all
three cases, as confirmed by both aggregate
performance metrics and agreement- and bootstrap-based analyses. Therefore, our general recommendation is to use these
modern CNN architectures for prediction models operating on PPG data.}
{CNN architectures such as XResNet1d consistently provided strong and stable performance
across all three cases. Therefore, our general recommendation is to consider CNN-based
architectures as a reliable baseline for prediction models operating on PPG data,
particularly under practical computational constraints and limited hyperparameter tuning.}
However, in certain cases, it may not be necessary to use such complex
models (e.g.\ see the Calibfree scenario of BP regression). Similarly,
for certain tasks, models leveraging image representations achieved
competitive results. Nevertheless, in the interest of achieving
\replaced{competitive results on an unknown task, the use of modern CNN
architectures based on raw time series representations remains the
safest choice}
{robust performance on an unknown task, CNN architectures operating on raw time-series
representations constitute a safe and effective baseline}.
In spite of promising results
\replaced{in other applications, more complicated hybrid CNN-RNN or
transformer-based models did not show any improvements over CNN-based
baselines}
{reported in other application domains, more complex hybrid CNN--RNN or
transformer-based models did not show consistent improvements over CNN-based
baselines in our experimental setup}.
This might not be a finding that generalizes to other datasets, but
potentially just a reflection of the fact that the considered models for
the respective datasets/tasks are already operating in a regime where
they are largely data-constrained and already close to the optimal
performance achievable on the respective datasets.

\textbf{Model Interpretability.} 
Beyond quantitative performance metrics, model interpretability represents a critical factor determining clinical adoption of AI systems. This interpretability challenge manifests differently across modelling approaches.
Feature-based methods offer inherent interpretability advantages through clinically meaningful inputs such as pulse morphology metrics or rhythm irregularity characteristics. These inputs provide immediate physiological interpretation that clinicians can readily understand. However, when complex prediction algorithms operate on these interpretable features, the overall model often becomes opaque, negating the initial interpretability benefit.

The explainable AI (XAI) community has developed techniques to address this opacity across different input representations. Post-hoc XAI methods, applicable to already-trained models, commonly generate heatmaps that highlight input regions contributing positively or negatively to model decisions \cite{holzinger2020explainable}. These approaches have shown promise when applied to physiological time series analysis \cite{wagner2024explaining}. Despite advances in XAI techniques, deep learning models operating on raw data remain fundamentally more challenging to interpret than models built on clinically meaningful features. This interpretability gap persists even with sophisticated explanation methods, creating a fundamental trade-off between model sophistication and clinical transparency.

\textbf{Robustness.} There is a second important quality dimension that is not captured explicitly by our analysis: robustness to input noise or artefacts. Note that wrist-worn PPG sensors, as used in DeepBeat, are more prone to motion artefacts and provide a lower signal-to-noise ratio relative to clinical-grade systems. But also within a single dataset, signal quality varies widely. \added{In free-living conditions, motion artefacts often dominate wrist PPG
recordings and represent one of the primary reasons why many academic PPG-based
models fail to translate into reliable real-world devices.} Future work should evaluate the robustness, for example, by considering predictive performance stratified according to signal quality. \added{While datasets such as DeepBeat or MIMIC would, in principle, allow such
analyses, the present study focuses on controlled, standardized evaluation
settings to enable like-for-like comparison of model architectures, and therefore
does not include an explicit quantitative robustness assessment.}  A second robustness limitation relates to
the restriction of assessing the performance purely based on
in-distribution performance, which is known to lead to an overly
optimistic assessment of the generalization performance. This urges for
dedicated studies on the out-of-distribution generalization performance
of the presented models, see \cite{ref60,ref59} for the first studies in the context of BP prediction.

\textbf{(Self-supervised) Pretraining.} An important restriction of the
presented study is the restriction to models that were trained from
scratch. A very promising extension lies in the consideration of
supervised or self-supervised pretraining, which, in the case of DeepBeat,
leads to an increase from 0.71 to 0.91 in terms of F1-scores for the
multi-task model. In particular, this applies to the recently published
foundation models for PPG data \cite{ref48,ref58}. 

\hypertarget{summary-and-conclusion}{%
\section{Summary and Conclusion}\label{summary-and-conclusion}}

Our investigations across two exemplary regression and classification
tasks found that in general modern CNNs (of the ResNet- or
Inception-kind) represented the best-performing approaches. The
competitiveness of small-scale models and feature-based approaches
depends crucially on the definition of the task at hand. While in some
scenarios (e.g.\ regression CalibFree), they can compete with or in some
cases even outperform modern CNNs, they may fail to show competitive
performance in other cases (e.g.\ regression Calib, classification). 

\subsection*{Code availability}
The complete implementation, including scripts for data preprocessing as
well as model training is available at
\url{https://gitlab.com/qumphy/d1-code}.

\subsection*{Generative AI statement}
No generative AI was used for text generation in this manuscript.

\subsection*{Competing interests}
The authors have declared no competing interests.

\subsection*{Ethics statement}
This study used publicly available and deidentified datasets. No direct patient interaction or intervention was involved. As these datasets are released under established data use agreements and have been ethically approved for secondary research, no additional ethics approval was sought.

\subsection*{Acknowledgments}
The project (22HLT01 QUMPHY) has received funding from the European
Partnership on Metrology, co-financed from the European Union's Horizon
Europe Research and Innovation Programme and by the Participating
States. Funding for the University of Cambridge, KCL, NPL and the University of Surrey 
was provided by Innovate UK under the Horizon Europe Guarantee Extension, 
grant numbers 10091955, 10087011, 10084125, 10084961 respectively.
PHC acknowledges funding from the British Heart Foundation (BHF) grant
{[}FS/20/20/34626{]}.

\bibliographystyle{IEEEtran}
\bibliography{bibfile}
\clearpage
\appendix
\hypertarget{supplementary-material}{%
\section{Supplementary material}\label{supplementary-material}}

\hypertarget{background}{%
\subsection{Background}\label{background}}

\hypertarget{blood-pressure-estimation-2}{%
\subsubsection{BP
estimation}\label{blood-pressure-estimation-2}}

BP is one of the most widely used physiological
measurements. It is a key marker of cardiovascular health; a valuable
predictor of cardiovascular events; and is essential for the selection
and monitoring of antihypertensive (BP lowering) treatments.
However, there are several limitations to BP measurements
currently taken in a clinical setting. First, clinical BP
measurements are not taken frequently in the wider population, meaning
many people around the world are not aware that they have high blood
pressure. Second, clinical BPs can be unrepresentative of a
patient's normal BP due to the white coat hypertension
effect, where a patient's BP is higher in the clinical
setting than normal life. Third, they provide only a snapshot of blood
pressure at a single time point, and do not capture diurnal variations
which contain important information on health. Fourth, cuff-based blood
pressure devices are not suitable for long-term monitoring.
Consequently, it would be highly valuable to develop technology to
monitor fluctuations in BP, unobtrusively in daily life,
such as in wearables.

PPG-based BP estimation provides a
potential approach to monitor BP unobtrusively in daily
life. PPG is now widely incorporated into consumer
wearable devices, and PPG signals vary with BP. Researchers
have focused on two main approaches to estimating BP from
PPG signals \footnote{Charlton PH \emph{et al.}, `Assessing hemodynamics
  from the photoplethysmogram to gain insights into vascular age: a
  review from VascAgeNet', \emph{American Journal of Physiology-Heart
  and Circulatory Physiology}, \emph{322}(4), H493--H522, 2022.
  \url{https://doi.org/10.1152/ajpheart.00392.2021}}: (i) using the
shape of a single PPG pulse wave; and (ii) using the speed at which the
pulse wave propagates through the arteries (the pulse wave velocity). In
this project, we focus on the first approach, since it can be
implemented in many common wearables such as smartwatches, whereas the
second requires simultaneous measurements at different anatomical sites.
BP can potentially be estimated from the shape of the pulse
wave using traditional signal processing based on expert-identified
features, or using deep learning techniques which learn how the pulse
wave is affected by BP changes for themselves.

\hypertarget{atrial-fibrillation-detection-2}{%
\subsubsection{AF
detection}\label{atrial-fibrillation-detection-2}}

AF is the most common sustained cardiac arrhythmia
and confers a five-fold increase in stroke risk. Fortunately, the risk
of stroke can be greatly reduced via anticoagulation. However, patients
with AF are often underdiagnosed, either because AF episodes occur
asymptomatically or because it occurs only intermittently and so is not
identified during routine testing. Therefore, there is a need for
continuous unobtrusive heart rhythm monitoring to identify patients with
AF.

PPG provides an attractive approach to identifying AF
because it is widely used in consumer wearables, and because it provides
a measure of the heart rhythm, which is irregular during AF. Indeed,
several smartwatches used PPG to identify irregular
rhythms which may be indicative of AF\footnote{Perez, M. V. \emph{et
  al.} (2019). Large-scale assessment of a smartwatch to identify atrial
  fibrillation. \emph{New England Journal of Medicine}, \emph{381}(20),
  1909--1917. \url{https://doi.org/10.1056/NEJMoa1901183}}. Irregular
heart rhythms have been identified from PPG signals by identifying
individual pulse waves corresponding to heartbeats, extracting
inter-beat intervals, and assessing the level of inter-beat interval
variability, where higher levels of variability indicate higher levels
of irregularity. More recently, deep learning models have been developed
to identify AF from the PPG signal \footnote{Ding, C. et al. (2024).
  PPG-based AF detection: A
  continually growing field. \emph{Physiological Measurement},
  \emph{45}(4), 04TR01. \url{https://doi.org/10.1088/1361-6579/ad37ee}}.

\hypertarget{original-deepbeat-dataset}{%
\subsection{Original DeepBeat dataset}\label{original-deepbeat-dataset}}

This dataset includes over 500,000 signal segments, each with a duration
of 25 seconds and a sampling rate of 32 Hz, from 175 individuals,
including 108 with AF and 67 without AF (Table \ref{table_6}).
The small test set and the overlapping signal segments of the original
dataset affected the analysis. This redundancy led to inflated
performance metrics, while the unbalanced distribution of AF and non-AF
subjects across training, validation and test sets further biased the
results. Consequently, despite its impressive performance on the test
set, the original model showed reduced effectiveness on validation and
training data. Finally, it is important to note key limitations of this
dataset for our study, all signal segments classified as AF were
collected from one study protocol where patients came in for a clinical
procedure. In contrast, all signal segments classified as non-AF were
collected from an independent study where subjects were doing an
exercise stress test. Therefore, different devices, human subject
parameters (comorbidities, heart rate) would have likely influenced the
classification performance, independently of AF. Therefore, confirmatory
studies on independent datasets are required before any conclusions
about AF classification can be drawn.

\begin{longtable}[]{@{}
  >{\raggedright\arraybackslash}p{(\columnwidth - 12\tabcolsep) * \real{0.1428}}
  >{\raggedright\arraybackslash}p{(\columnwidth - 12\tabcolsep) * \real{0.1429}}
  >{\raggedright\arraybackslash}p{(\columnwidth - 12\tabcolsep) * \real{0.1429}}
  >{\raggedright\arraybackslash}p{(\columnwidth - 12\tabcolsep) * \real{0.1429}}
  >{\raggedright\arraybackslash}p{(\columnwidth - 12\tabcolsep) * \real{0.1429}}
  >{\raggedright\arraybackslash}p{(\columnwidth - 12\tabcolsep) * \real{0.1428}}
  >{\raggedright\arraybackslash}p{(\columnwidth - 12\tabcolsep) * \real{0.1428}}@{}}

\toprule\noalign{}
\begin{minipage}[b]{\linewidth}\raggedright
Set
\end{minipage} & \begin{minipage}[b]{\linewidth}\raggedright
Subjects
\end{minipage} & \begin{minipage}[b]{\linewidth}\raggedright
Total samples
\end{minipage} & \begin{minipage}[b]{\linewidth}\raggedright
AF samples
\end{minipage} & \begin{minipage}[b]{\linewidth}\raggedright
Non-AF samples
\end{minipage} & \begin{minipage}[b]{\linewidth}\raggedright
Data ratio
\end{minipage} & \begin{minipage}[b]{\linewidth}\raggedright
AF ratio
\end{minipage} \\
\midrule\noalign{}
\endhead
\bottomrule\noalign{}
\endlastfoot
Train & 137 & 2799784 & 1269660 & 1540124 & 0.839 & 0.45 \\
Validation & 16 & 518782 & 47706 & 471175 & 0.156 & 0.09 \\
Test & 22 & 17617 & 4230 & 13387 & 0.005 & 0.24 \\

\caption{Original DeepBeat (Stanford Wearable PPG) Dataset \label{table_6}} \\ 
\end{longtable}

\hypertarget{performance-evaluation-and-metrics-1}{%
\subsection{Performance evaluation and
metrics}\label{performance-evaluation-and-metrics-1}}

\textbf{BP regression} In our BP regression
task, we use the \(L_{1}\) norm to evaluate and compare the performance
of the machine learning models. Specifically, we employ the empirical
equivalent of this norm, known as the \emph{Mean Absolute Error}
(MAE)\emph{,} defined as
follows:
\begin{equation}
MAE\, = \,\frac{1}{N}\sum_{i = 1}^{N}\left| y_{i} - \widehat{y_{i}} \right|,
\end{equation}
where \(y_{i}\) denotes the ground truth values, and \(\widehat{y_{i}}\)
the predictions made by the machine learning model on the test set.

\textbf{AF classification} Detection performance is evaluated by
counting the correctly detected AF cases (true positives, \(N_{TP}\)),
correctly detected non-AF cases (true negatives, \(N_{TN}\)), falsely
detected AF cases (false positives, \(N_{FP}\)), and missed AF cases
(false negatives, \(N_{FN}\)). Here, the term "case" refers to a
segment. Using these four counts, several commonly used performance
metrics are calculated.

\emph{Sensitivity} (Se) describes the probability of a positive result
given that the sample is truly positive, i.e.
\begin{equation}
Se\, = \,\frac{N_{TP}}{N_{TP} + N_{FN}}.
\end{equation}
\emph{Specificity} (Sp) describes the probability of a negative result
given that the sample is truly negative, i.e.
\begin{equation}
Sp\, = \,\frac{N_{TN}}{N_{TN} + N_{FP}}.
\end{equation}
The \emph{receiver operating characteristic} (ROC) curve illustrates the
performance of a binary classifier at varying threshold values by
depicting the rate of truly classified positives against the rate of
falsely classified positives. The area \emph{under the curve} (AUC) of
the ROC indicates the overall model performance by integrating the ROC
over the threshold. The AUC ranges from 0 to 1, with 1 representing
perfect detection of positives and negatives and 0 representing perfect
misclassification. Moreover, a value of 0.5 represents random
classification of the samples. We note that although AUC is widely used,
it integrates sensitivity and specificity across both relevant and
irrelevant clinical regions and is sensitive to dataset
imbalance\footnote{Lobo, J. M., Jiménez-Valverde, A., \& Real, R.
  (2008). AUC: a misleading measure of the performance of predictive
  distribution models. Global Ecology and Biogeography, 17(2), 145-151.}\textsuperscript{,}\footnote{Hanczar,
  B., Hua, J., Sima, C., Weinstein, J., Bittner, M., \& Dougherty, E. R.
  (2010). Small-sample precision of ROC-related estimates.
  Bioinformatics, 26(6), 822-830.}. Therefore, it is most useful to
understand model performance under different parameter settings during
training rather than report general model performance\footnote{Butkuvienė,
  M., Petrėnas, A., Sološenko, A., Martín-Yebra, A., Marozas, V., \&
  Sörnmo, L. (2021). Considerations on performance evaluation of atrial
  fibrillation detectors. IEEE Transactions on Biomedical Engineering,
  68(11), 3250-3260.}.

The F1 score describes the harmonic mean of the \emph{precision}, i.e.,
the rate of true positives against the number of all positive labels,
and the \emph{recall}, i.e., the rate of true positives against the
number of all truly positive samples and describes the overall
performance of a classifier. The F1 score is defined via
\begin{equation}
F_{1} = \frac{2\, N_{TP}}{2\, N_{TP}\, + \, N_{FP}\, + \, N_{FN}}.
\end{equation}

The F1 score ranges from 0 to 1, with 1 indicating perfect detection and
0.5 indicating performance equivalent to random detection.

The Matthew's correlation coefficient (MCC) measures the difference
between the predicted values of a classifier and the actual class
values. MCC is equivalent to the \(\chi^{2}\)-statistics for a 2 x 2
contingency table. The MCC is defined via
\begin{equation}
MCC\, = \,\frac{N_{TP}N_{TN}\, - \, N_{FP}N_{FN}}{\sqrt{\left( N_{TP} + N_{FP} \right)\left( N_{TP} + N_{FN} \right)\left( N_{TN} + N_{FP} \right)\left( N_{TN} + N_{FN} \right)}}.
\end{equation}
In the original form, MCC ranges from -1 to 1, where 1 indicates perfect
detection and -1 indicates complete inverse detection. For easier
comparison, MCC is typically normalized\footnote{Chicco, D., \& Jurman,
  G. (2020). The advantages of the Matthews correlation coefficient
  (MCC) over F1 score and accuracy in binary classification evaluation.
  BMC genomics, 21, 1-13.}, however, note that MCC, in contrast to the
F1 score, accounts for the number of correctly classified true
negatives. MCC is especially useful for imbalanced datasets, as it
reflects performance when most AF and non-AF episodes are accurately
detected.

In binary classification, the model outputs a raw class confidence value
between 0 and 1. However, most evaluation metrics require a binary
classified value as input. To classify a sample using the confidence
value, we set a classification threshold. If the confidence is below
this threshold, the sample is assigned to the negative class; otherwise,
it is assigned to the positive class. The threshold is an arbitrary
parameter that can be set to achieve the best performance.

For the F1 score, we set the threshold to the naive value of 0.5. The
AUC score uses the raw confidence value and does not require a
threshold. For the remaining metrics, we choose the threshold that
achieves a sensitivity (specificity) on the test set that is closest to,
but greater than, 0.8. Once this threshold is set, we calculate the
remaining metrics and denote them accordingly.

\hypertarget{methods}{%
\subsection{Methods}\label{methods}}

\hypertarget{raw-time-series-models}{%
\subsubsection{Raw Time Series Models}\label{raw-time-series-models}}

Researchers have investigated a variety of deep learning models for the
prediction of BP using PPG and
electrocardiogram (ECG) signals. For instance, a study by Kachuee et al.
(2017)\footnote{Kachuee, M., Kiani, M. M., Mohammadzade, H., \& Shabany,
  M. (2017). Cuffless BP estimation algorithms for
  continuous health-care monitoring. IEEE Transactions on Biomedical
  Engineering, 64(4), 859-869.} developed a deep neural network model to
estimate BP using ECG and PPG signals, which showed
significant improvements in accuracy and reliability compared to
traditional methods.

Another notable study by Liang et al. (2018)\footnote{Liang, Y., Chen,
  Z., Ward, R., \& Elgendi, M. (2018). Hypertension assessment via ECG
  and PPG signals: An evaluation using machine learning models.
  Computational and Mathematical Methods in Medicine, 2018, Article
  3179780} introduced a hybrid model that integrates CNN and recurrent
RNN to exploit both spatial and temporal features of PPG and ECG
signals. This model exhibited enhanced performance in continuous blood
pressure monitoring, highlighting the potential of deep learning in
medical signal processing. It is worth mentioning that transfer learning
has been applied to improve the performance of BP prediction
models. Zhang et al. (2020)\footnote{Zhang, Y., Feng, X., Li, J., Li,
  B., \& Peng, X. (2020). Transfer learning for hybrid deep neural
  network-based BP estimation. Journal of Healthcare
  Engineering, 2020, Article 5462540.} utilized a pre-trained deep
learning model and fine-tuned it with a smaller dataset of PPG and ECG
signals. This approach minimized the requirement for extensive labelled
data, enhancing the practicality of the model for real-world
applications.

In this project, we employ prominent models such as
LeNet1d\footnote{LeCun, Y., Bottou, L., Bengio, Y., \& Haffner, P.
  (1998). Gradient-based learning applied to document recognition.
  Proceedings of the IEEE, 86(11), 2278-2324.}, Inception1d\footnote{Szegedy,
  C., Liu, W., Jia, Y., Sermanet, P., Reed, S., Anguelov, D., ... \&
  Rabinovich, A. (2015). Going deeper with convolutions. In Proceedings
  of the IEEE conference on computer vision and pattern recognition (pp.
  1-9).}, XResNet50d, XResNet101d\footnote{Strodthoff, Nils, Temesgen
  Mehari, Claudia Nagel, Philip J. Aston, Ashish Sundar, Claus Graff,
  Jørgen K. Kanters et al. "PTB-XL+, a comprehensive
  electrocardiographic feature dataset." \emph{Scientific data} 10, no.
  1 (2023): 279.}, AlexNet1d\footnote{Krizhevsky, Alex, Ilya Sutskever,
  and Geoffrey E. Hinton. "Imagenet classification with deep
  convolutional neural networks." Advances in neural information
  processing systems 25 (2012).} and Minirocket\footnote{Dempster,
  Angus, Daniel F. Schmidt, and Geoffrey I. Webb. "Minirocket: A very
  fast (almost) deterministic transform for time series classification."
  Proceedings of the 27th ACM SIGKDD conference on knowledge discovery
  \& data mining. 2021.} as well as Temporal Convolutional Networks, to
conduct a thorough assessment of the outcomes.

\hypertarget{baseline}{%
\paragraph{Baseline}\label{baseline}}

The baseline model predicts BP by using the median value of
the BP data, providing a straightforward reference for
evaluating the performance of more advanced predictive models.

\hypertarget{xresnetd50-101}{%
\paragraph{XResNetd50 /101}\label{xresnetd50-101}}

XResNet50d and XResNet101d represent deep learning state-of-the-art
models derived from ResNet architecture. They incorporate more layers,
50 and 101 layers, respectively, with a number of other enhancements,
including group normalization and the option for selective kernel sizes.
This allows these models to learn even more intricate hierarchical
features and to be very powerful in tasks dealing with complex
time-series data. The GNLL variant featured two parallel output branches
and dropout layers dispersed throughout the architecture. Separate
models were trained for SBP and DBP prediction for each dataset. Each
input was evaluated 100 times with dropout layers left active, and the
prediction is the mean of the output means.

\hypertarget{inception1d}{%
\paragraph{Inception1d}\label{inception1d}}

Inspired by the Inception architecture but for one-dimensional signals,
it uses parallel convolutional layers with different kernel sizes. This
model captures information at multiple scales, enabling the model to
understand %\st{much} 
complex patterns in time series better.

\hypertarget{lenet1d}{%
\paragraph{\texorpdfstring{LeNet1d }{LeNet1d }}\label{lenet1d}}

LeNet1 is a one-dimensional convolutional neural network adapted from
the original LeNet architecture, which is optimized for time-series data
analysis. Considering a greater number of convolutional and pooling
layers, it is more appropriate for extracting features from univariate
signals, such as PPG or ECG.

\hypertarget{alexnet1d}{%
\paragraph{AlexNet1d}\label{alexnet1d}}

The AlexNet1d is a 1-dimensional adaptation of the popular convolutional
neural network (CNN) architecture, traditionally used for 2D image
classification. This model consists of two primary components:

\begin{enumerate}
\def\labelenumi{\arabic{enumi}.}
\item
  Feature Extractor: A stacked series of convolutional, ReLU (Rectified
  Linear Unit), and max pooling layers.
\item
  Classifier: A fully connected neural network comprising two linear
  layers, each followed by a ReLU activation function and a dropout
  layer, to prevent overfitting. The final layer is a linear layer that
  outputs the model\textquotesingle s predictions.
\end{enumerate}

During training, we use the Gaussian negative log likelihood as the cost
function, allowing the model to learn the mean and standard deviation of
the predictive probability density.

\hypertarget{minirocket}{%
\paragraph{MiniRocket}\label{minirocket}}

The MiniRocket is an almost deterministic transform for time-series
data, followed by a linear classifier. This approach uses dilated
convolutions with kernels of length 9 and all possible permutations of
the weights -1 and 2 that sum up to 0, where the dilation values are
selected to be within a fixed range relative to the input length.
Following the convolution, features are extracted by calculating the
proportion of positive values in each convolution output, which is the
number of positive values, divided by the sequence length. This process
yields a total of 9,996 features, which are then fed into a linear
regressor to predict, in our case, the mean and standard deviation of a
Gaussian predictive probability density.

The kernels do not require training, allowing for a one-time feature
extraction. This enables efficient model training, as only the linear
regressor needs to be trained over multiple epochs. This makes the
MiniRocket a very fast and effective baseline for time-series
classification and regression tasks.

\paragraph{PPNet} The PPNet incorporates CNNs and LSTMs units to deal with physiological time series. CNNs characterize local spatial patterns in biosignals, and LSTMs are responsible for handling temporal dependencies, ultimately creating an efficient model to differentiate complex biomedical patterns.
\paragraph{iTransformer} iTransformer turns the classical transformer on its head by applying self-attention across features instead of time steps. Each variable is treated as a token, and the model can look at inter-variable interactions. Instance-normalization specificity also preserves signal characteristics, and there are improved results on multivariate time-series forecasting tasks.
\paragraph{TimesNet}
TimesNet adopts an inception-style convolution architecture specific to time series. Through parallel convolutions with different receptive fields, it characterizes multi-periodic patterns in different scales in time.

\hypertarget{temporal-convolutional-networks-tcns}{%
\paragraph{Temporal Convolutional Networks
(TCNs)}\label{temporal-convolutional-networks-tcns}}

Temporal Convolutional Networks (TCNs) are a class of neural networks
designed for sequential data processing. Their ability to capture both
short-term and long-term dependencies makes them particularly effective
for analyzing complex biological time series signals, such as
electrocardiograms (ECGs), electroencephalograms (EEGs), and other
physiological measurements. TCNs have been used for both forecasting and
classification tasks\footnote{Lin, Y., Koprinska, I., \& Rana, M.
  (2020). Temporal Convolutional Neural Networks for Solar Power
  Forecasting, 2020 International Joint Conference on Neural Networks
  (IJCNN), Glasgow, UK, 2020, pp. 1-8, doi:
  10.1109/IJCNN48605.2020.9206991.},\footnote{Pelletier, C., Webb, G.
  I., \& Petitjean, F. (2019). Temporal convolutional neural network for
  the classification of satellite image time series. Remote Sensing,
  11(5), 523.},\footnote{Hewage, P., Behera, A., Trovati, M., Pereira,
  E., Ghahremani, M., Palmieri, F., \& Liu, Y. (2020). Temporal
  convolutional neural (TCN) network for an effective weather
  forecasting using time-series data from the local weather station.
  Soft Computing, 24, 16453-16482.} yet, to the best of our knowledge,
TCNs have never been used for the analysis of PPG signals while we
believe TCNs can be relevant for such a task. Indeed, TCNs leverage the
strengths of convolutional neural networks (CNNs) and adapt them for
temporal data. Unlike recurrent neural networks (RNNs) that process data
sequentially, TCNs use convolutional layers to capture temporal
dependencies, allowing for parallel processing and typically achieving
faster training times. TCNs main features are:

\begin{itemize}
\item
  Causal Convolutions: They ensure that predictions at time t are only
  influenced by inputs from time t and earlier to maintain the temporal
  order.
\end{itemize}

\[y(t) = \sum_{i = 0}^{k - 1}w_{i}x(t - i)\]

\begin{quote}
where \(x(t)\) and \(y(t)\) are respectively the input and output at
time t, \(w_{i}\) are the weights of the convolution filter, and \(k\ \)
stands for the filter size. This equation ensures that the output
\(y(t)\) depends only on the current and previous inputs, preserving the
causality.
\end{quote}

\begin{itemize}
\item
  Dilated Convolutions: They allow to capture long-range dependencies
  using a large receptive field without increasing computational load.
\end{itemize}

\[y(t) = \sum_{i = 0}^{k - 1}w_{i}x(t - d\, \cdot \, i)\]

\begin{quote}
where \(d\ \) represents the dilation factor (which determines the spacing
between the elements of the input signal). Dilated convolutions allow
the network to cover a larger range of inputs without increasing the
number of parameters. Indeed, a dilated convolution lets the network look
back up to $dk$ time steps, enabling exponentially larger receptive fields
per the number of layers. In the original paper, the authors increased
\(d\) exponentially with the depth of the network.
\end{quote}

\begin{itemize}
\item
  Residual Connections: They help to mitigate the vanishing gradient problem and
  improve the training of deeper networks.
\end{itemize}

\hypertarget{clinically-interpretable-feature-cif-based-models}{%
\subsubsection{Clinically Interpretable Feature (CIF)-Based
Models}\label{clinically-interpretable-feature-cif-based-models}}

\hypertarget{preprocessing}{%
\paragraph{Preprocessing}\label{preprocessing}}

Raw PPG signals often show reduced quality due to motion artifacts,
baseline drift, and hypoperfusion. Motion artifacts result from body
movement or improper sensor attachment, introducing signal fluctuations
that degrade signal quality. Baseline drift, caused by respiration and
body movements, shifts the PPG waveform and may obscure the true
pulsatile component. Hypoperfusion, characterized by reduced peripheral
blood flow due to vasoconstriction, weakens PPG signals, affecting the
accuracy and reliability of physiological measurements.

While CNN-based deep learning networks develop custom filter banks
during training and require no specific preprocessing, expert-crafted
feature-based methods necessitate signal preparation. Typically, the
lower bound of the passband (\textasciitilde0.5 Hz) removes components
below 0.1 Hz and respiratory influences in the 0.1--0.5~Hz range,
preserving the AC component. The upper bound (\textasciitilde5--10 Hz)
retains primary PPG frequency components, often captured using
Butterworth, Chebyshev II, or finite impulse response (FIR) filters.

In this study, for BP estimation, PPG segments from the
VitalDB dataset were preprocessed using a Butterworth infinite impulse
response zero-phase band-pass filter (4th order, 0.4-7 Hz). For atrial
fibrillation detection, PPG segments from the DeepBeat dataset were
preprocessed using low-pass, high-pass, and adaptive filters.
High-frequency noise and artifacts were removed using a low-pass
infinite impulse response filter (cutoff 6 Hz), while baseline wander
was eliminated using a high-pass infinite impulse response filter
(cutoff 0.5 Hz) and a fifth-order normalized least mean squares adaptive
filter with a reference input of unity.

\hypertarget{feature-extraction-for-bp-estimation}{%
\paragraph{Feature extraction for BP
estimation}\label{feature-extraction-for-bp-estimation}}

\hypertarget{clinically-interpretable-features}{%
\paragraph{Clinically Interpretable
Features}\label{clinically-interpretable-features}}

\begin{figure}[t]  
  \centering
  \includegraphics[width=\columnwidth]{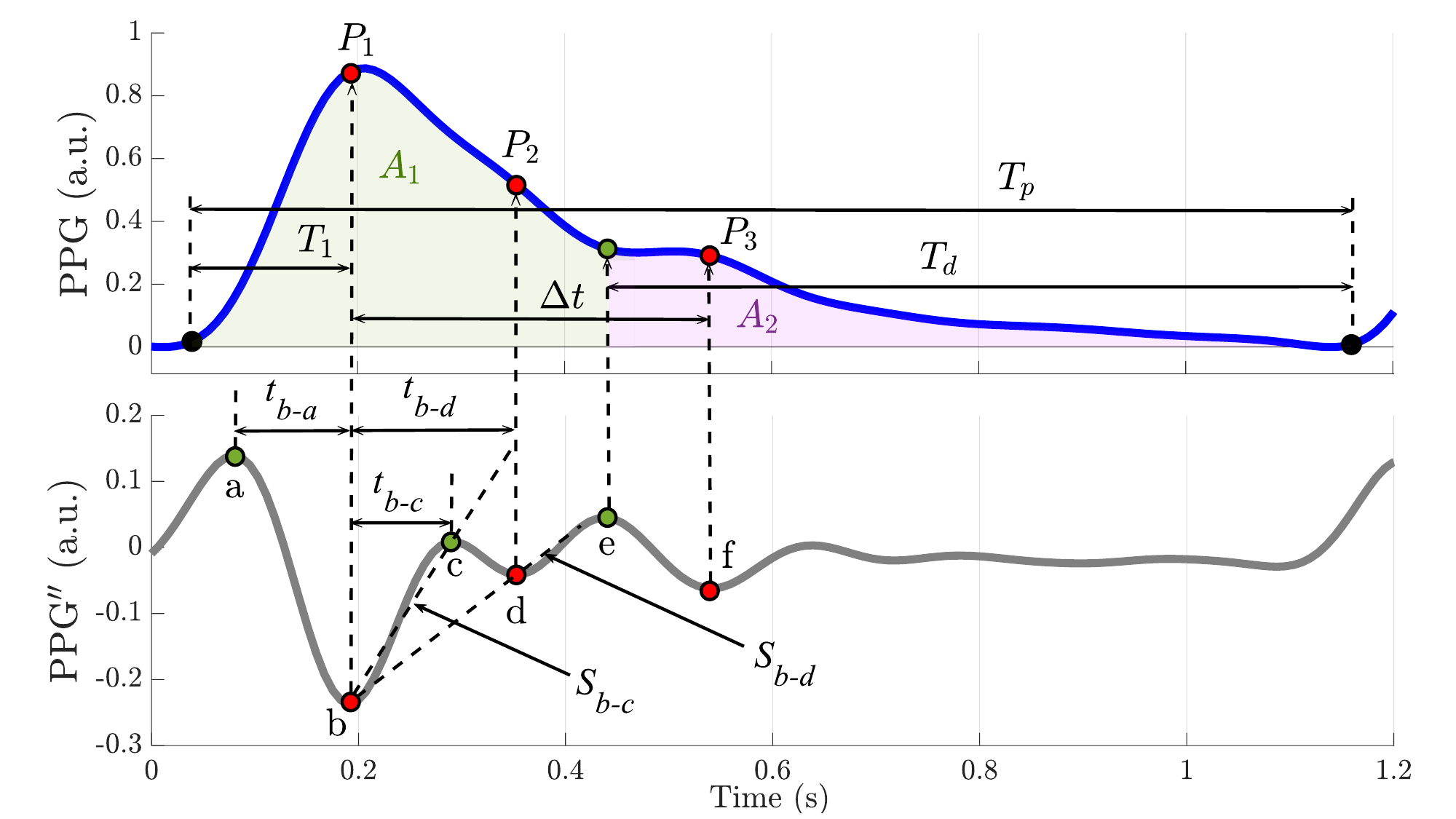}
  \caption{The definition of PPG pulse wave morphology features. Higher-order statistical features extracted from PPG pulse waveforms
include skewness and kurtosis, with the latter being the most
statistically significant feature for blood pressure estimation.} 
  \label{fig:fig2}
\end{figure}

To estimate BP, 28 PPG features\footnote{Charlton, P.
  H., Celka, P., Farukh, B., Chowienczyk, P., \& Alastruey, J. (2018).
  Assessing mental stress from the photoplethysmogram: a numerical
  study. Physiological measurement, 39(5), 054001.} were assessed based
on pulse morphology analysis and pulse derivative features (Figure \ref{fig:fig2}),
which are features selected from the literature\textsuperscript{24} as the
most significant for BP estimation.
  
PPG pulse wave features include amplitude-related parameters such as the 
\added{amplitudes of the first (\emph{P\textsubscript{1}}), second (\emph{P\textsubscript{2}}) and third (\emph{P\textsubscript{3}}) pulse waveforms\footnote{Kontaxis, S., Gil, E., Marozas, V. et al. (2021) Photoplethysmographic waveform analysis for autonomic reactivity assessment in depression. IEEE Trans. Biomed. Engng. 68(4), 1273--1281},}
\deleted{first (\emph{P\textsubscript{1}}) and second (\emph{P\textsubscript{2}})
systolic peaks, the diastolic peak (\emph{P\textsubscript{3}}),} and
derived indices like the \emph{P\textsubscript{2}/P\textsubscript{1}}
ratio, reflection index (\emph{RI =
P\textsubscript{3}/P\textsubscript{1}}), and augmentation index
(\emph{AI = (P\textsubscript{1} --
P\textsubscript{3})/P\textsubscript{1}}). Time-related features consist
of the pulse duration (\emph{T\textsubscript{p}}), diastolic duration
(\emph{T\textsubscript{d}}), systolic duration
(\emph{T\textsubscript{1}}), and the time interval from
\emph{P\textsubscript{1}} to \emph{P\textsubscript{3} ($\Delta$t)}.
Area-related features include the systolic area
(\emph{A\textsubscript{1}}), diastolic area (\emph{A\textsubscript{2}}),
the inflection point area ratio (\emph{IPA =
A\textsubscript{2}/A\textsubscript{1}}), and the inflection point area
plus the d-wave amplitude of the second PPG derivative (\emph{IPAD}).

PPG derivative features include amplitude-related parameters such as
ratios of the second PPG derivative waveform amplitudes (\emph{b/a, c/a,
d/a,} and \emph{e/a}), the cardiovascular aging index (\emph{AGI =
(b-c-d-e)/a}) for arterial stiffness assessment, the interval aging
index (\emph{AGI\textsubscript{int} = (b-e)/a}), and the modified aging
index (\emph{AGI\textsubscript{mod} = (b-c-d)/a}). Time-related features
include time intervals between second PPG derivative waves
(\emph{t\textsubscript{b-a}, t\textsubscript{b-c},
t\textsubscript{b-d}}). Slope-related features include the slope
coefficients (\emph{slope\textsubscript{b-c}, slope\textsubscript{b-d}})
of straight lines between amplitudes of \emph{b} and \emph{c}, and
\emph{b} and \emph{d} waves, respectively.

\hypertarget{quasi-periodic-signal-features}{%
\paragraph{(Quasi)-periodic signal
features}\label{quasi-periodic-signal-features}}

To solve the BP regression problem, we employ a Wavelet
Packet Decomposition (WPD) approach using the Discrete Wavelet Transform
(DWT) for feature extraction from raw PPG-signals. Specifically, we
utilize the Daubechies wavelet (db6) of order 3, which offers a balance
between time and frequency localization, making it particularly suitable
for analysing non-stationary signals.

The wavelet packet decomposition allows for a hierarchical
representation of the signal by decomposing both the approximation and
detail the coefficients at each level. This provides a finer level of signal
analysis compared to standard DWT, capturing more detailed information
across different frequency bands.

After obtaining the wavelet packet coefficients, these features are
flattened and then directly fed, without additional processing, as inputs
to an MLP model. This simple approach will also
be used to diagnose AF from PPG-signals.

\hypertarget{feature-extraction-for-af-detection}{%
\paragraph{Feature extraction for AF
detection}\label{feature-extraction-for-af-detection}}

\hypertarget{irregularity-features}{%
\paragraph{Irregularity features}\label{irregularity-features}}

\begin{figure}[t]  
  \centering
  \includegraphics[width=\columnwidth]{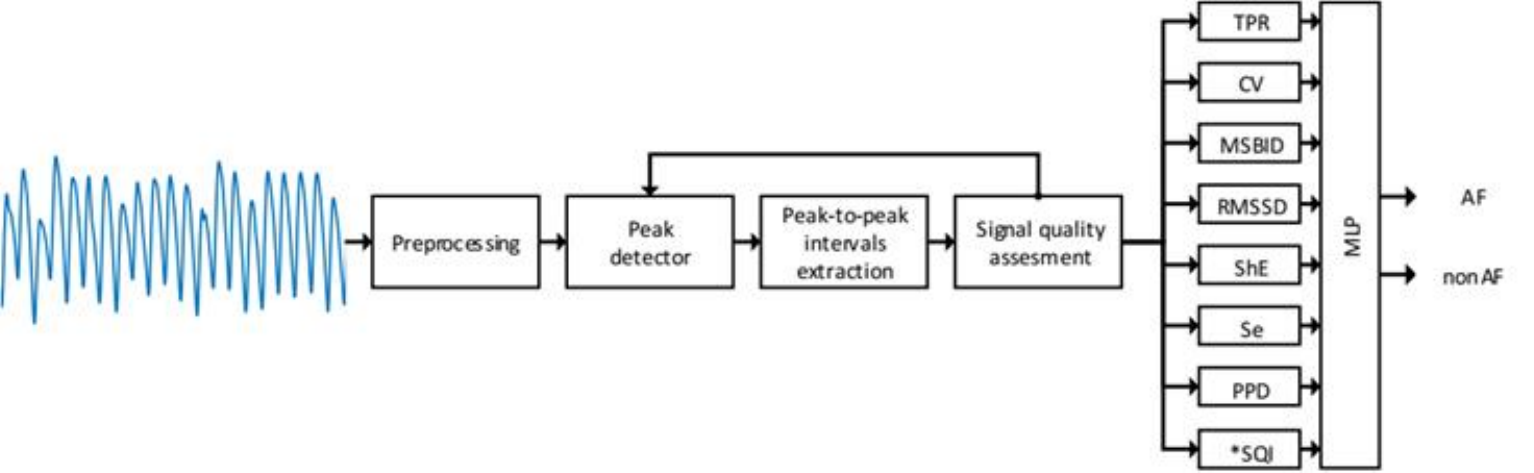}
  \caption{Structure of an MLP-based model for the AF detection task using
features extracted from the PPG signal.} 
  \label{fig:fig3}
\end{figure}

The irregularities in peak-to-peak (PP) intervals extracted from PPG
signals can be evaluated based on measures of randomness, variability,
and complexity\footnote{Petrėnas, A., \& Marozas, V. (2018). Atrial
  fibrillation from an engineering perspective (pp. 137-220). L. Sörnmo
  (Ed.). Berlin: Springer.}. The structure of an MLP-based atrial
fibrillation detection algorithm using clinically interpretable features
extracted from the PPG signal is shown in Figure \ref{fig:fig3}.

\begin{figure}[t]  
  \centering
  \includegraphics[width=\columnwidth]{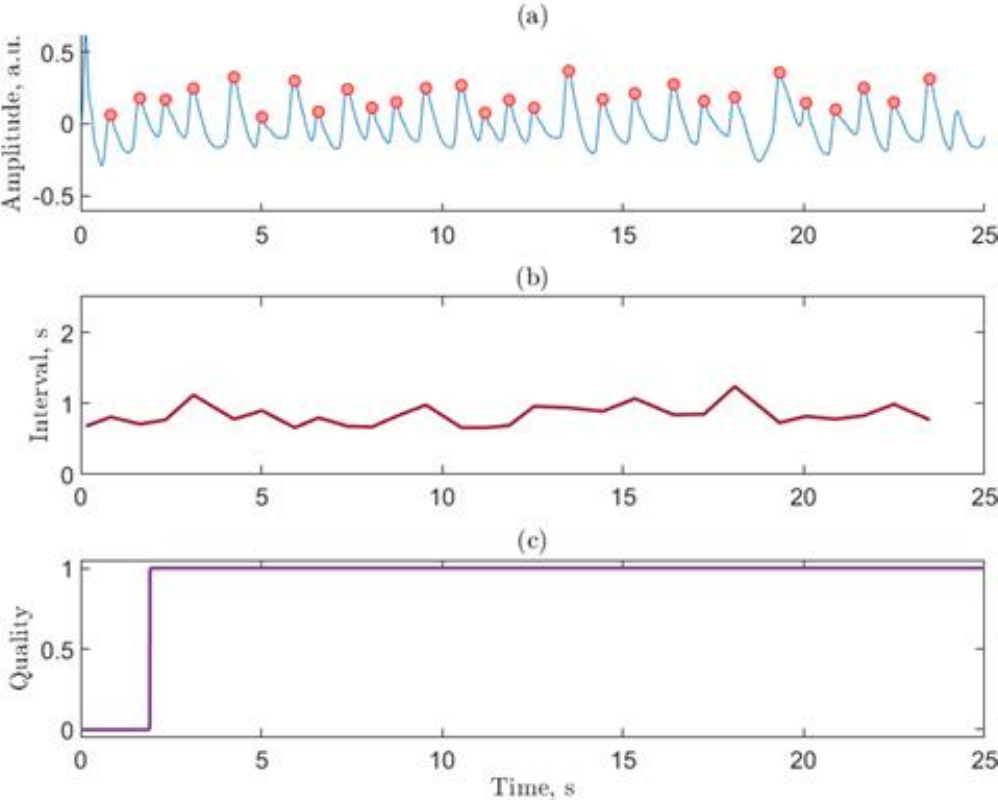}
  \caption{Good-quality PPG segment for the AF task (a) with extracted PP intervals (b)
and quality index (c).} 
  \label{fig:fig4}
\end{figure}

\begin{figure}[t]  
  \centering
  \includegraphics[width=\columnwidth]{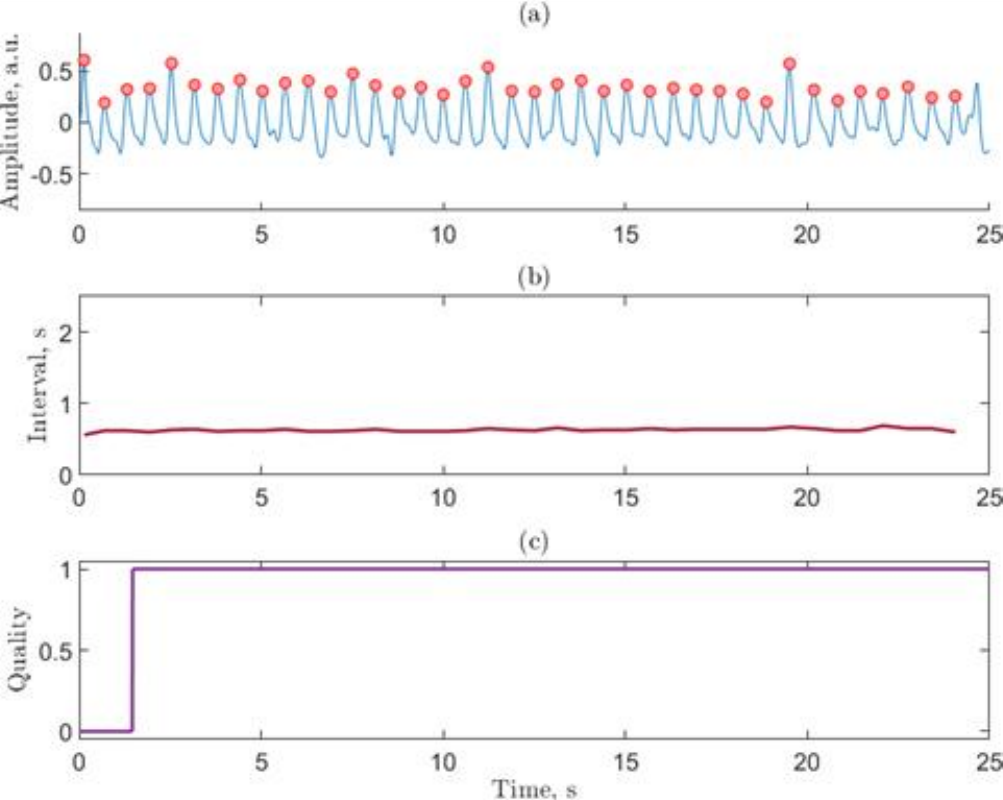}
  \caption{Good-quality PPG segment for the AF task (a) with extracted PP intervals (b)
and quality index (c).} 
  \label{fig:fig5}
\end{figure}

\begin{figure}[t]  
  \centering
  \includegraphics[width=\columnwidth]{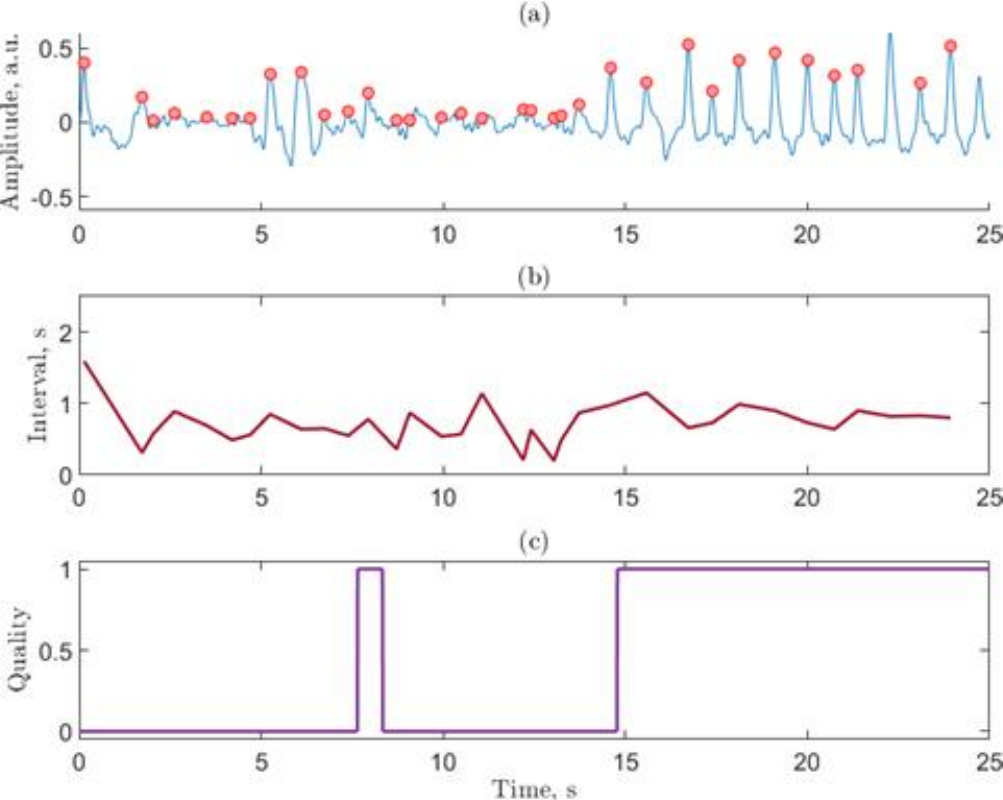}
  \caption{Bad-quality PPG segment for the AF task (a) with extracted PP intervals (b)
and quality index (c).} 
  \label{fig:fig6}
\end{figure}

\begin{figure}[t]  
  \centering
  \includegraphics[width=\columnwidth]{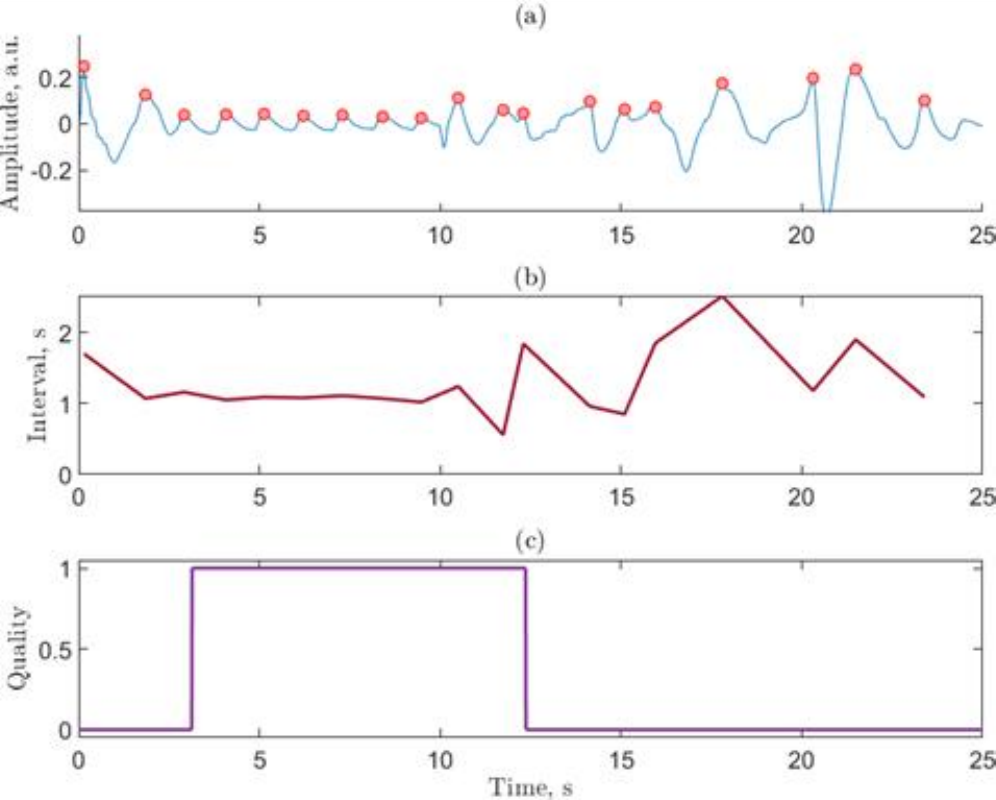}
  \caption{Bad-quality PPG segment for the AF task (a) with extracted PP intervals (b)
and quality index (c).} 
  \label{fig:fig7}
\end{figure}

Figure \ref{fig:fig4} to Figure \ref{fig:fig7} show PP intervals extracted from 25-s duration PPG
segments which are of good quality with AF, good quality without AF, bad
quality with AF, and bad quality without AF, respectively.

The following irregularity features are used:

\begin{itemize}
\item
  Turning point ratio (TPR)\footnote{Dash, S., Chon, K. H., Lu, S., \&
    Raeder, E. A. (2009). Automatic real-time detection of atrial
    fibrillation. Annals of biomedical engineering, 37, 1701-1709.}
  evaluates randomness in PP intervals by identifying turning
  points---intervals greater or less than their two neighbours. It is
  computed as the ratio of turning points to total PP intervals within
  the analysis window. A higher TPR indicates greater randomness, aiding
  AF detection.
\item
  Coefficient of variation (CV)\footnote{Tateno, K., \& Glass, L.
    (2001). Automatic detection of AF using the
    coefficient of variation and density histograms of RR and $\Delta$RR
    intervals. Medical and Biological Engineering and Computing, 39,
    664-671.}\textsuperscript{,}\footnote{Langley, P., Dewhurst, M., Di
    Marco, L. Y., Adams, P., Dewhurst, F., Mwita, J. C., ... \& Murray,
    A. (2012). Accuracy of algorithms for detection of atrial
    fibrillation from short duration beat interval recordings. Medical
    engineering \& physics, 34(10), 1441-1447.} relates the standard
  deviation of PP intervals to their mean. In AF, increased dispersion
  and decreased mean duration raise CV.
\item
  Mean successive beat interval difference (MSBID)\footnote{Langley, P.,
    Dewhurst, M., Di Marco, L. Y., Adams, P., Dewhurst, F., Mwita, J.
    C., ... \& Murray, A. (2012). Accuracy of algorithms for detection
    of atrial fibrillation from short duration beat interval recordings.
    Medical engineering \& physics, 34(10), 1441-1447.} also links PP
  interval dispersion to the mean, similar to CV.
\item
  Root mean square of successive differences (RMSSD)\textsuperscript{51}
  measures dispersion without considering mean heart rate. It typically
  increases during AF.
\item
  Shannon entropy (ShE)\footnote{Shannon, C. E. (1948). A mathematical
    theory of communication. The Bell system technical journal, 27(3),
    379-423.} quantifies signal unpredictability, rising when
  distribution is uniform and dropping when centered. ShE is often
  higher in AF than sinus rhythm.\footnote{Zhou, X., Ding, H., Ung, B.,
    Pickwell-MacPherson, E., \& Zhang, Y. (2014). Automatic online
    detection of atrial fibrillation based on symbolic dynamics and
    Shannon entropy. Biomedical engineering online, 13(1), 1-18.}.
\item
  Sample entropy (SE)\footnote{Richman, J. S., \& Moorman, J. R. (2000).
    Physiological time-series analysis using approximate entropy and
    sample entropy. American journal of physiology-heart and circulatory
    physiology, 278(6), H2039-H2049.} assesses self-similarity by
  calculating the probability that patterns persist in extended samples.
  A simplified SE is used here for efficient AF detection.\footnote{Petrėnas,
    A., Marozas, V., \& Sörnmo, L. (2015). Low-complexity detection of
    atrial fibrillation in continuous long-term monitoring. Computers in
    biology and medicine, 65, 184-191.}.
\item
  Poincare plot (PPD)\footnote{Park, J., Lee, S., \& Jeon, M. (2009).
    Atrial fibrillation detection by heart rate variability in Poincare
    plot. Biomedical engineering online, 8(1), 1-12.}\textsuperscript{,}\footnote{Lian,
    J., Wang, L., \& Muessig, D. (2011). A simple method to detect
    atrial fibrillation using RR intervals. The American journal of
    cardiology, 107(10), 1494-1497.} visualizes consecutive PP
  intervals. In AF, it shows significantly greater dispersion compared
  to sinus rhythm or ectopic beats.
\end{itemize}

\begin{figure}[t]  
  \centering
  \includegraphics[width=\columnwidth]{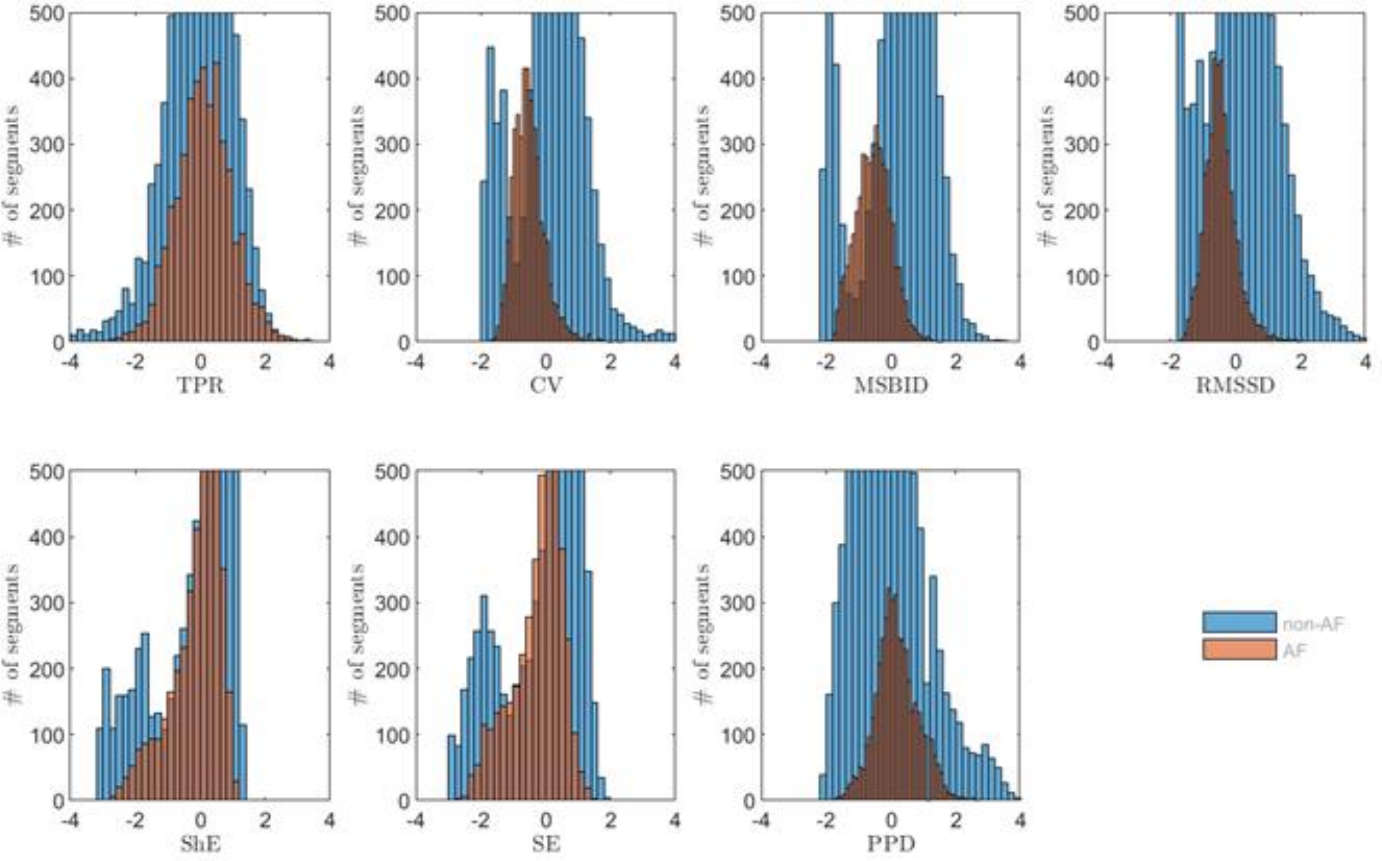}
  \caption{Discriminative capacity of each feature to distinguish between
AF and non-AF cases without removing bad-quality PPG segments.} 
  \label{fig:fig8}
\end{figure}

\begin{figure}[t]  
  \centering
  \includegraphics[width=\columnwidth]{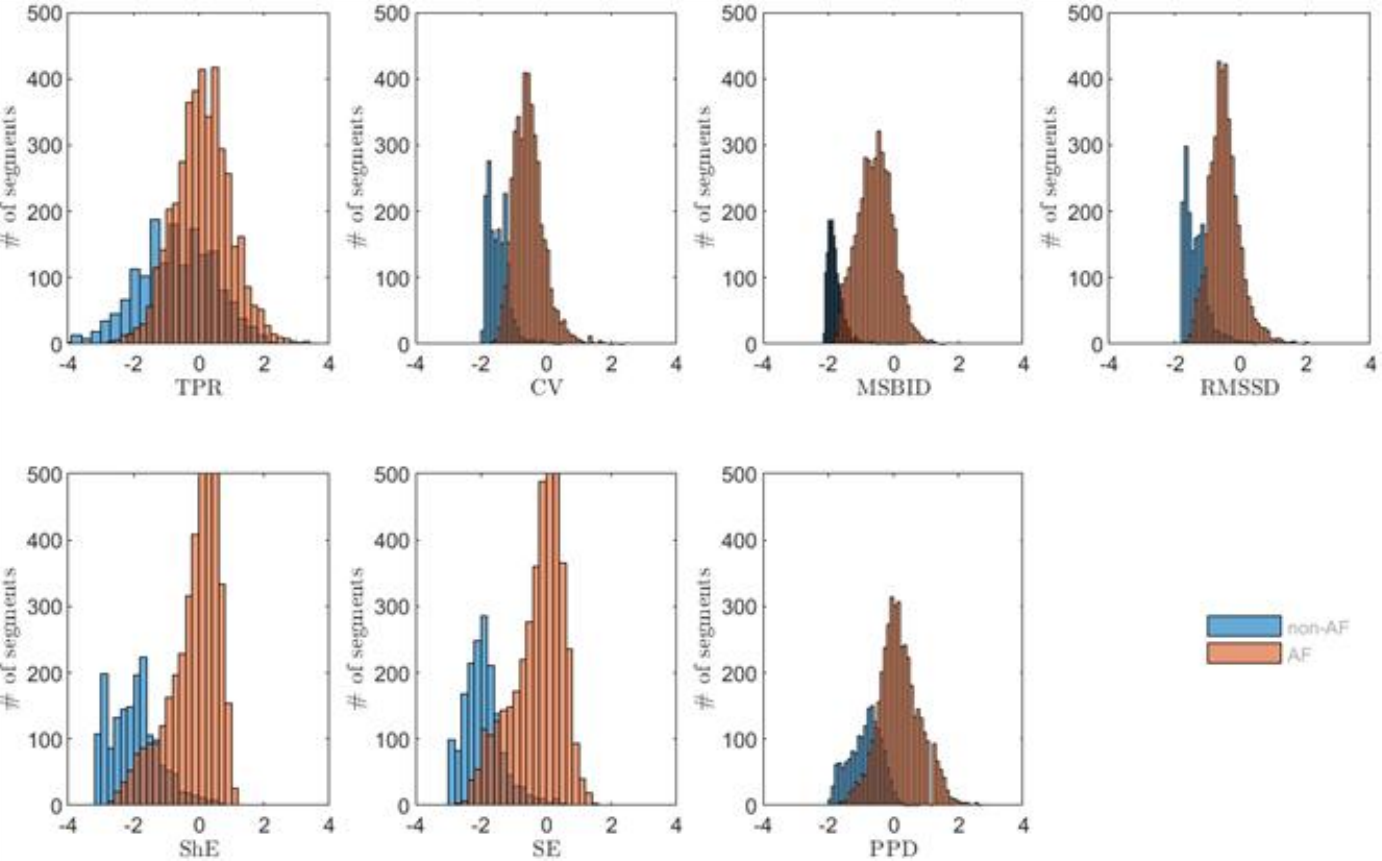}
  \caption{Discriminative capacity of each feature to distinguish between
AF and non-AF cases using the removal of bad-quality PPG segments.} 
  \label{fig:fig9}
\end{figure}

The capacity of each feature to discriminate between AF and non-AF cases
is depicted in Figure \ref{fig:fig8} and Figure \ref{fig:fig9}. Figure \ref{fig:fig8} shows separation of
features before the removal of bad-quality PPG segments, while Figure \ref{fig:fig9}
shows separation of features after the removal of bad-quality PPG
segments. PPG signal quality assessment was performed using the
algorithm described in\footnote{Sološenko, A., Petrėnas, A., Paliakaitė,
  B., Sörnmo, L., \& Marozas, V. (2019). Detection of atrial
  fibrillation using a wrist-worn device. \emph{Physiological
  measurement}, \emph{40}(2), 025003.}. The best visible separation of
features is achieved by using SE and PPD, especially when estimated
bad-quality segments are removed.

\hypertarget{models}{%
\paragraph{Models}\label{models}}

\hypertarget{multilayer-perceptron-models}{%
\paragraph{Multilayer perceptron
models}\label{multilayer-perceptron-models}}

The multilayer perceptron (MLP) used for classifying features extracted
from PPG signals consists of a fully connected layer with 128 neurons,
followed by a ReLU activation function and a dropout layer with a 0.5
dropout rate for regularization. The network\textquotesingle s output
structure includes a fully connected layer with 2 neurons, a softmax
layer to produce a probability distribution for multi-class
classification, and a classification layer for determining the final
class label. The model is trained using the Adam optimization algorithm,
with fixed learning rate of 0.01 and mini-batch size of 32.

\hypertarget{gaussian-process-regression}{%
\paragraph{Gaussian process
regression}\label{gaussian-process-regression}}

Gaussian Process Regression (GPR) is a powerful and flexible
non-parametric regression technique used in machine learning and
statistics for modelling complex, nonlinear relationships\footnote{Rasmussen,
  C. E. and C. K. I. Williams. Gaussian Processes for Machine Learning.
  MIT Press. Cambridge, Massachusetts, 2006.}. Instead of fitting a
specific function to the data, GPR models the relationship between input
features and output as a distribution over functions. It incorporates
prior knowledge (expressed through kernels) and provides uncertainty
estimates for individual predictions. The objective of GPR is to find a
mean function that accurately represents observed data points and can be
used for predicting new data points. Gaussian Processes define a
distribution over an infinite number of potential functions that could
fit the data.

The key points of the GPR method include the use of the Fully
Independent Conditional (FIC) approximation for making predictions given
the model parameters and the squared exponential kernel as the
covariance function. The inter-point distance computation for evaluating
built-in kernel functions was specified as the training function
$(x-y)^2$. To ensure accurate parameter
estimation, a QR-factorization-based approach was employed for computing
the log-likelihood and gradient. Additionally, the limited-memory
Broyden-Fletcher-Goldfarb-Shanno (LBFGS) optimizer was used for
hyperparameter estimation. As a quasi-Newton method, LBFGS approximates
the Hessian matrix using past gradient information, enabling rapid
convergence in high-dimensional parameter spaces while maintaining
computational feasibility and numerical stability. The combination of
QR-factorization, the squared exponential kernel, a linear basis
function, and the FIC approximation enhances the predictive performance
and scalability of the model.

\hypertarget{image-based-models}{%
\subsubsection{Image-Based Models}\label{image-based-models}}

\hypertarget{continuous-wavelet-transform-cwt-image-representations}{%
\paragraph{Continuous Wavelet Transform (CWT) image
representations}\label{continuous-wavelet-transform-cwt-image-representations}}

CWT transformation requires to define several hyperparameters among
which the most important are the (1) the choice of the wavelet function,
(2) the number of scales and, (3) the window size for the wavelet
function at each scale. Indeed, scales affect the frequency resolution,
higher scales focusing on lower frequencies and longer time windows. The
number of scales defines how many different frequency bands the CWT will
compute. In this project, we use a generalized morse wavelet (GMW) as
the mother wavelet with a length of 1024 and 128 logarithmically spaced
scales. Applying the CWT transform to PPG signals with shape
(\textless sequence\_length\textgreater, 1) yields a 2D array with
size ((\textless sequence\_length\textgreater, 128) which is saved as an
RGB image.

State-of-the art methods relying on CWT based scalogram as inputs, feed the CWT images to a deep learning model to perform
classification/regression tasks. We propose to use the same model (ResNet18) for both BP estimation and AF
detection, the only difference being the addition of a sigmoid function for
AF detection.

The only input preprocessing step simply consists in resizing CWT images into
squared RGB images with shape (224,224,3). Such input size is typically
dictated by the availability of pre-trained models commonly trained on the
ImageNet dataset.

\hypertarget{STFT-Based Spectrogram Representation (Spectrogram-ResNet)}{%
\paragraph{STFT-Based Spectrogram Representation (Spectrogram-ResNet)}\label{Spectrogram-ResNet}}

\added{We represent PPG signals using fixed resolution time-frequency maps based on the STFT. This was done to enable modeling using convolutional neural networks. This process is placed inside the preprocessing level. Every input signal is formulated in the form of a $(T, F)$ sequence, where $T$ denotes the number of temporal samples and $F$ the number of frequency bins. Moreover, to decrease the effect of boundaries on the computation of the STFT, symmetric padding is applied to both ends of the signal in the time-domain. This resulted in a channel-wise representation of power spectrograms with uniform temporal and frequency resolution. The resulting representations were organized as $(N_{\text{windows}}, F, T)$ and subsequently converted to the logarithmic domain using decibel scaling to improve numerical stability and dynamic range. These log-power spectrograms work as input to the trained ResNet-50 or ResNet-18 CNN. The proposed Spectrogram-ResNet approach is capable of processing hierarchical information of the PPG signal in the time-frequency representation, complementing the currently existing time-domain-based models.}

\added{
We represent PPG signals using fixed-resolution time--frequency maps based on the short-time Fourier transform (STFT) to enable modeling with convolutional neural networks. This processing is performed at the preprocessing stage. Each input signal is initially represented as a one-dimensional time series of length $T$. The STFT transforms this signal into a time--frequency representation of size $(T, F)$, where $T$ denotes the number of temporal frames and $F$ the number of frequency bins. 

To reduce boundary effects during STFT computation, symmetric padding is applied to both ends of the signal in the time domain. This results in power spectrograms with uniform temporal and frequency resolution. The resulting representations are organized as $(N_{\text{windows}}, F, T)$ and subsequently converted to the logarithmic domain using decibel scaling to improve numerical stability and dynamic range. These log-power spectrograms serve as input to the trained ResNet-50 CNN. 

The proposed Spectrogram-ResNet approach captures hierarchical patterns in the PPG signal through its time--frequency representation, complementing existing time-domain-based models.
}

\hypertarget{Models ParametersCount}{%
\subsubsection{Models Parameters Count}\label{Models-Parameters-Count}}

\begin{table}[ht]
\centering
\begin{tabular}{@{}llcc@{}}
\toprule
\textbf{Type} & \textbf{Model} & \textbf{AF Detection (Classification)} & \textbf{BP Estimation (Regression)} \\
\midrule
 & XResNet1d101     & 1.8 M& 1.8 M\\
 & XResNet1d50      & 886 K& 886 K\\
 & Inception1d      & 472 K& 472 K\\
 & LeNet1d          & 429 K& 658 K\\
 & AlexNet1d        & 42 M& 56,7 M\\
T & MiniRocket       & 30 K& 30 K\\
 & iTransformer     & 642 K& 642 K\\
 & TimesNet         & 667 K& 667 K\\
 & PPNet            & 125 K& 125 K\\
 & TCN+MLP          & 6,7 M& 10,4 M\\
 & XResNet1d50+GNLL & 1,7 M& 1,7 M\\
\midrule
 & Wavelet+MLP      & 611 K& 840 K\\
F & CIF+MLP          & 1410 & between 4705 to 11882 \\
 & CIF+GPR          & --- & 32 \\
\midrule
 & CWT              & 11,1 M& 11,1 M\\
 I
& \added{Spectrogram-ResNet} & 23.5 M& 23,5 M\\
\bottomrule
\end{tabular}
\caption{Comparison of total parameter counts between models used for AF detection and BP estimation tasks, grouped by input representation type (\textbf{T}: time-series, \textbf{F}: feature-based, \textbf{I}: image-based, \textbf{B}: baseline).}
\label{table_param_counts}
\end{table}

\end{document}